\documentclass[12pt]{article}
\usepackage[utf8]{inputenc}

\usepackage[english]{babel}

\usepackage{bm}

\usepackage{fullpage}

\usepackage{amsmath}
\usepackage{amsfonts}

\usepackage{graphicx}
\usepackage[colorlinks=true, allcolors=blue]{hyperref}
\usepackage{subcaption}
\usepackage{natbib}
\usepackage[svgnames]{xcolor}
\usepackage[T1]{fontenc}
\usepackage{tcolorbox}
\usepackage{soul}
\usepackage{pdfpages}

\usepackage{tipa}
\DeclareFontFamilySubstitution{T3}{lmr}{cmr}

\newcommand{\dAIC}{\Delta\textrm{AIC}}

\title{Letters hide the truth from our eyes: \\ English homophones have meaningfully different phonetic realizations}
\author{
Yu-Hsiang Tseng$^{1}$,
Mirjam T. C. Ernestus$^{2}$,\\
Louis F. M. ten Bosch$^{2}$,
R. Harald Baayen$^{1}$\\[0.5em]
$^{1}$University of Tübingen, Germany\\
$^{2}$Radboud University, the Netherlands
}

\date{}
\begin{document}
\maketitle
\begin{abstract}
\noindent
The distribution of spoken word duration of English homophones is known to co-vary with frequency of use.  This study investigates whether other aspects of the phonetic realization of homophones also differ. A series of quantitative investigations of 14,000 homophone tokens in American television news broadcasts revealed that the tokens of homophone pairs such as \textit{weight} and \textit{wait} have different phonetic realizations, and that these can be predicted from their meanings in utterance context. These systematic differences remain even when taking duration-related variation into account. Time-normalized spectrograms emerged as an excellent tool for probing the fine details of phonetic realization, and obviate the need for phonetic transcriptions, which inevitably hide the phonetic truth from our eyes.
\end{abstract}

\newpage

\section{Introduction}

\noindent
Do homophones sound the same? \citet{Gahl2008} and \citet{Lohmann2018} provided evidence that this is not the case. A survey of the spoken word durations of English homophones in the Switchboard corpus \citep{godfrey1992switchboard} revealed that higher-frequency homophones (e.g. \textit{time}) tend to be shorter than the corresponding lower-frequency homophones (e.g., \textit{thyme}).  One possible reason that frequency-related durational differences arise is that their entries in the mental lexicon have their own frequency-based resting activation level. Other possible reasons are that higher-frequency words are more predictable in context and more likely to benefit from articulatory routinization.  

A different explanation was offered by \citet{Gahl2024}. This study showed that the spoken word duration of homophones could be predicted to a considerable extent from the meanings of these homophones. They showed that the greater the support a word form receives from its meaning, the longer the spoken word duration. 

This finding indicates that meaning and form are tightly linked, challenging formal phonology \citep[following in the footsteps of ][]{Port:Leary:2005}, as well as derivative models in psychology, according to which abstract phonological units shield articulation from semantics. This tight coupling between meaning and form also aligns with the extensive literature on phonesthemes, recurring sound–meaning pairings that are not readily analyzable as morphemes \citep{Bergen2004}. For example, English words beginning with ``gl-''  (e.g., \textit{glimmer}, \textit{glisten}, \textit{glitter}) are often associated with light or vision. Such non-arbitrary pairings have been consistently identified in quantitative studies across languages \citep[e.g.][]{Monaghan2014,Blasi2016,Pimentel2019}, and similar effects have also been observed in iconic correspondences between English spoken word forms and the sounds of their referents, such as the word \textit{frog} and frog croaks \citep{deVarda2025}. These systematicities suggest that members of a homophone pair, carrying different meanings, will not sound identical.

The present study asks whether, within each English heterographic \textit{homophone pair} (e.g. \textit{wait} and \textit{weight}), the two \textit{homophone members} differ not only with respect to the distributions of their spoken word durations, but also with respect to their phonetic realization. A further question considered in the present study is whether differences in phonetic realization can be traced back to differences in meaning.  In what follows, we show that the answer to both questions is yes.  

In section~\ref{sec:homophones}, we introduce key findings in the research in psychology on English homophones, and section~\ref{sec:phonetics} zooms in on the literature on phonetic realization differences between segmentally similar or identical words.  The theoretical framework adopted in the present study, the Discriminative Lexicon Model \citep{Heitmeier2025}, is introduced in section~\ref{sec:theory}. Section~\ref{sec:modeling} presents a series of quantitative investigations of the spectrograms of 14,000 tokens of 35 homophone pairs extracted from the Redhen resource \citep{Joo2017}.  These investigations provide strong support for the possibility that how a homophone token is articulated depends on its meaning in context.  Section~\ref{sec:word-duration} shows that the same patterns are observed after controlling for word duration. The implications of these findings for formal phonology and cognitive models of the mental lexicon are presented in the general discussion (Section~\ref{discussion}).

\newpage

\section{Homophones}\label{sec:homophones}

\noindent
Homophones are words that are said to sound the same but that have different meanings. Examples from English are heterographic homophones such as \textit{wait} and \textit{weight}, and homographic homophones such as financial \textit{bank} and river \textit{bank}.   

Homophones have been studied extensively as they offer the analyst the possibility to study meaning while controlling for form. In psychology, there are informed theories of the organization of the mental lexicon. Classical models of the mental lexicon typically posit multiple layers of representations, distinguishing between layers with syntactic, morphological, and phonological units. Nodes in these layers pass activation to one another, either in a cascading manner or through interactive feedback \citep[e.g.,][]{Dell1986,Roelofs1997,Levelt1999}. As homophones are words that have different meanings but the same pronunciation, they are assumed to share the same phonological representations. For example, \citet{Levelt2001} distinguished three strata: a conceptual stratum that accounts for conceptual semantics, a lemma stratum that accounts for words' morpho-syntactic properties, and a form stratum that contains units for stems and exponents. According to this model, a speaker first starts with a concept (e.g., \textsc{weight} or \textsc{wait}), then selects the corresponding lemma (which provides pointers to information about word category, N for \textit{weight} and V for \textit{wait}). The lemma in turn passes activation on to the shared phonological form \mbox{/}we$^{\mbox{\tiny i}}$t\mbox{/}.  The phonological form activates the corresponding phones, which are then bundled into syllables, which are passed on to the articulators. 

According to this model by \citet{Levelt2001}, the word frequency effect is located at the level of lexemes.\footnote{An alternative explanation of the word frequency effect in this model links frequency of use to a verification process.} If homophones share the same lexeme, then the frequency effect for high- and low-frequency homophonic words should be the same. \citet{Jescheniak1994} tested this prediction using experiments with Dutch-English bilinguals. They compared participants' reaction times in a translation task with those in a semantic judgment task, and found that low-frequency homophones with high-frequency counterparts behaved like non-homophonic high-frequency words. They argued that a less frequent word (e.g., \textit{nun}) ``inherits'' the frequency of its more frequent homophone twin (e.g., \textit{none}) through their shared lexeme representation. 

\citet{Dell1990} observed that not only do high-frequency words elicit fewer speech errors, but low-frequency words with high-frequency homophone twins are less prone to speech errors as well. This study explained this observation by again assuming that homophone twins have distinct \textit{lemma} nodes encoding word identity, but share the same \textit{lexeme} node encoding phonological form.  In Dell's model, interactive activation flowing between lemma and lexeme nodes leads to greater lexeme activation for low-frequency homophones than would be expected on the basis of their frequency alone. 

However, not all empirical findings align with the hypothesis that homophones share the same form representations. \citet{Caramazza2001}, using a picture naming task, compared three types of words: English low-frequency words  (e.g. \textit{nun}) with high-frequency homophones (e.g. \textit{none}), controls matched for word-specific frequency (e.g., \textit{owl}), and controls matched with cumulative-homophone frequency (e.g., \textit{tooth}, matching the combined frequency of \textit{none} and \textit{nun}). After further controlling for potential confounds between visual and articulatory complexity, they observed reaction times for homophones to be more similar to the reaction times to word-specific matched controls. They did not find any evidence for frequency inheritance. Further studies investigating different languages, different kinds of homophones, and using different tasks also failed to find consistent evidence for shared form representations for homophones \citep{Miozzo2005,Miozzo2003,Jescheniak2003}. 

A further problem for the hypothesis of homophones having a shared form representation was reported by \citet{Ferreira2003}. They suggested that lemma selection may not be entirely encapsulated from phonological processes. In their experiments, participants were asked to name pictures (e.g., of a \textit{priest}). Before the picture was presented, participants first read a cloze sentence that was completed either with a semantic competitor of the target object name (e.g., \textit{nun}), a homophone of the competitor (e.g., \textit{none}), or an unrelated control (e.g., \textit{match}). Interestingly, picture naming showed interference not only from semantically related words, but also from semantically unrelated homophones. This interference is unexpected in models with a strict top-down architecture, but may be compatible with interactive activation architectures. \citet{Ferreira2003} and \citet{Miozzo2005} argue for partially independent form representations and feedback between semantics and phonological information, suggesting a closer interaction between form and meaning in homophones.

\section{The phonetic realization of homophones}\label{sec:phonetics}

If homophones share the same form representation driving articulation, then the prediction is that there are no systematic differences in the phonetic realization of homophones \citep{Kiparsky1982,Dell1986,Levelt1999}.

However, \citet{Gahl2008} observed for heterographic homophones in English that more frequent homophones (e.g., \textit{time}) in a pair tend to have shorter spoken word duration in spontaneous speech compared to their lower-frequency homophone twins (e.g., \textit{thyme}), after controlling for variables such as speech rate, predictability, syntactic category and orthographic regularity. \citet{Lohmann2018} reported that duration also varies as a function of syntactic category: words such as \textit{cut} have different spoken durations when used as nouns as compared to verbs.

Another example of words that were originally supposed to be homophones but, upon closer inspection, turned out to differ in their phonetic realization are words with syllable-final voiced and unvoiced obstruents in German and Dutch. Syllable-final obstruents have been described as being realized as voiceless \citep{Kenstowicz1994}, resulting in homophone pairs such as German \textit{Rad} ``wheel'' and \textit{Rat} ``council'', or Dutch \textit{ik verwijd} ``I widen'' and \textit{ik verwijt} ``I reproach'' \citep{Port1985,Warner2004}. Although these pairs should be phonologically identical under the rule of final devoicing, a range of studies nevertheless found that the voice neutralization in these word pairs is incomplete.  The word-final de-voiced obstruents are realized differently from their unvoiced counterparts with respect to vowel duration, burst duration, and the number of glottal pulses. While the effect of the incomplete neutralization is small and may involve knowledge of the written word forms \citep{Fourakis1984,Warner2006}, empirical findings show that not only do speakers produce the obstruent differently, but that listeners can also perceive the subtly different duration cues \citep{Port1989,Ernestus2006,Roettger2014}.

Affixes have also been reported to be realized with different pronunciation durations, depending on their morphological status. For English, \citet{Plag2017} observed that the duration of word-final `s' is longer when it is non-morphemic as compared to when it is an inflectional exponent or a clitic.  Furthermore, the /s/ as plural exponent tends to be longer than the clitic /s/.  These findings dovetail well with the study of \citet{Drager2011}, which reported for conversational English that the grammatical and discourse functions of \textit{like} co-determine the realization of pitch and the degree of diphthongization.


Finally, the investigation of \citet{Seyfarth2017} indicated that morphologically distinct homophones differ phonetically.  Because they have different morphological relatives, fricative-final inflected words (e.g. \textit{laps}) are argued to have longer stem and suffix duration than their morphologically simple counterparts (e.g., \textit{lapse}), even though both share the same phones.  These results suggest that the phonetic implementation of segmentally identical morphologically complex words is shaped by their morphological structure \citep{Song2013,Plag2017,Celata2022,Schmitz2024}, but see \citet{saito2024articulatory} for an alternative explanation.  A discussion of the role of re-syllabification and concomitant prosodic restructuring of otherwise segmentally identical stems in Dutch and English is provided by \citet{Kemps2005a,Kemps2005b}.

In summary, there is more systematic variation in the phonetic realization of segmentally similar or identical words. This systematicity has been argued to be due to differences in frequency of use, differences in word category, differences in morphological structure, and semantic transparency. The abovementioned study by \citet{Drager2011} on English ``like'' adds word meaning to the set of explanatory factors.  

Using methods from distributional semantics and computational modeling (which we discuss in more detail below), \citet{Gahl2024} showed for English homophones that word meaning is a strong predictor of spoken word duration. For Mandarin single-syllable homophones, \citet{Jin:Ernestus:Baayen:2026} observed differences in the details of the phonetic realization of their tones, indicating that pitch is a further dimension in which differences in meaning between words sharing the same segments can be expressed. Similar form-meaning alignment has also been reported for Mandarin tone at the word token level \citep[see, e.g.,][]{Chuang:Bell:Tseng:Baayen:2025,lu2025realization} and also for the realization of pitch in English left-stressed two-syllable words \citep{chuangwords}.

The goal of the present study is to clarify whether fine-grained high-dimensional context-specific representations of the meanings of word tokens of homophones are not only aligned with spoken word duration, but also with other aspects of the fine details of phonetic realization of homophones. Building on the observation of \citet{Drager2011} that how segments are realized phonetically depends on words' meanings, in what follows we investigate whether consistent differences in the phonetic realization of homophones exist, and whether these differences can be predicted, at least in part, from their meanings.

\section{Theoretical and methodological framework}\label{sec:theory}

The present research is theoretically anchored in the Discriminative Lexicon Model \citep[DLM;][]{baayen2019discriminative,Heitmeier2025}, a computational model of the mental lexicon that implements mappings between forms and meanings to model language comprehension and production. These mappings predict that the details of words' meanings unavoidably give rise to differences in phonetic realization, not only for spoken word duration and the realization of F0 contours, but also for how words are articulated \citep[for an articulography study, see][]{saito2024articulatory}. The mappings are typically implemented with networks that constitute both the model's memory and its processing algorithm. The model does not work with underlying forms, and it does not have entries for words' forms or meanings.


In the DLM, a form-and-meaning mapping is learned by continually adjusting the association weights between inputs and outputs. For discrete inputs (for comprehension, following \citet{Danks2003}, referred to as cues) and for discrete outputs (referred to as outcomes), the mechanism by which the associations are updated can be elegantly described by a simple learning rule, in which the weights are strengthened when a cue and an outcome co-occur, and weakened when a cue is present but an outcome is absent \citep{Rescorla1972}. For example, seeing a large flat surface (cue) of a grand piano (outcome) will strengthen their association, but weaken the association between a large surface and a table. Experimental evidence shows that seeing a table followed by a grand piano in a naming experiment slows naming responses \citep{Marsolek2008}. This suggests that the association weights of discriminative cues are continually recalibrated by real-life exposure to co-occurring events in the environment. For continuous inputs and outputs, the learning rule of \citet{Widrow1960} follows the same underlying learning mechanism, with an update rule that iteratively reduces prediction error.

There are no static representations of forms and meanings in DLM; only the association weights are stored for comprehension and production, and these are continuously updated and recalibrated with experience. For comprehension, a word's form is an ephemeral input to the network. The network subsequently calculates a corresponding meaning, which is also not stored. After a mapping is completed, the network's weights are updated.  For production, the process of conceptualization is assumed to result in an ephemeral semantic representation that is input to the network mapping meaning to form.  Once the form has been estimated, the weights of the production network are updated. 

These form-meaning mappings of the DLM are mostly implemented as simple linear transformations, not unlike multivariate linear regression. If so required, non-linearity can be added to increase prediction accuracy \citep{Chuang:Bell:Tseng:Baayen:2025,Heitmeier2025}. Using simple linear mappings has the advantage that no intermediate layers of hidden nodes are involved that filter the information flowing between form and meaning, and that may not be well interpretable linguistically.  Moreover, by using linear mappings, it becomes possible to investigate to what extent semantic detail and phonetic realization co-vary systematically. 

The form and meaning representations that the DLM works with consist of lists of numbers. In the discrete binary case, each number indicates whether a cue or outcome is present; in the continuous real-valued case, it indicates how strongly a cue or outcome is activated.  When many cues and outcomes are involved, such lists can be treated as high-dimensional vectors. A vector is commonly described as an object with a direction and magnitude, and most intuitively visualized as an oriented arrow in a 2-D or 3-D space. The high dimensionality of embeddings makes it possible to capture wide arrays of highly nuanced similarities and differences. In what follows, we introduce the form and meaning vectors that are central to the present study.

\subsection{Representing Meanings}\label{sec:repr_meanings}

Vectors in the meaning space of the DLM are referred to as \textit{semantic vectors}. One of the widely adopted semantic vectors in computational linguistics is the so-called word vectors, also known as word embeddings, which have their roots in distributional semantics \citep{Harris1954,Firth1957}. Although word embeddings are often associated with natural language processing \citep{Mikolov2013,Pennington2014,Bojanowski2017}, the idea of representing meanings as vectors has a long tradition also in the psycholinguistic literature \citep{Lund1996,Landauer1997,Shaoul2010,Griffiths2007}. The embeddings referenced thus far are all type-based, that is, one word --- in the sense of a letter string --- is represented by one vector, regardless of the context the word occurs in. For example, the orthographic form \textit{bank} in ``financial bank'' and ``river bank'' is associated with the very same embedding.  

To avoid the limitations of type-specific embeddings, in the present study, we make use of contextualized embeddings (CEs). These token-based embeddings represent the meaning of a word differently depending on the context in which it occurs.  CEs usually are vectors with greater dimensionality as they require more capacity to differentially represent lexical meanings across contexts. Early studies on CEs trained a simple recurrent network and showed that the same word types have different CEs if they are embedded in different sentence structures \citep{Elman1991,Elman2004}. Later studies and models have pursued many different architectures \citep[e.g.,][]{Bengio2003,Jones2007,Sutskever2011,Radford2018,Peters2018,Peng2023,Dao2024}.

For the present study, we extracted contextualized embeddings at the word token level from a large language model (LLM; here, GPT-2). This model has 137M parameters and a tokenizer with 50,257 unique string and substring tokens. Both model and tokenizer are available at \url{https://huggingface.co/openai-community/gpt2}.\footnote{    
The model tokenized the texts before computing embeddings. Orthographic words are mapped to one or more `model tokens', which can be short words or substrings of longer words. For example, \textit{see} and \textit{seen} are both associated with one `model token' each, but \textit{capitol} is tokenized into ``cap'' and ``-itol''.}

The homophones used in the present study are all tokenized into one `model token',  \textit{capitol} being the only exception. For this word,  we averaged the embeddings of both `model tokens' to obtain the CE. For each model token, the input text consisted of a five-word window immediately preceding (and including) the target word. The context-specific meaning of a word token is represented as a 768-dimensional vector. For studies assessing the quality and potential of CEs, see \citet{Manning2020,Linzen2021,Tenney2019,Pavlick2022,Schrimpf2021,Caucheteux2022,Goldstein2024,Chuang:Bell:Tseng:Baayen:2025,lu2025realization}.

\subsection{Representing Forms}
\label{speech-form}

Previous research investigating the phonetic properties of spoken words has studied, for example, spoken word duration \citep{Hay2004,Gahl2008,Seyfarth2017}, spectral-temporal measurements \citep{Port1989}, and time-series of f0 measurements \citep{Chuang:Bell:Tseng:Baayen:2025,lu2025realization}. In the present study, instead of extracting partial information from the speech signal, we focus on the spectrogram of that speech signal.

Our point of departure is a word token's Mel-scale spectrogram, which is an efficient compressed representation that is informed by the human cochlear frequency-place map \citep{Stevens1937,Greenwood1990}. Current deep learning-based speech models extract feature representations from MEL spectrograms \citep[e.g.,][]{Baevski2020,Radford2023}. These feature representations have been successfully applied in linguistic studies modeling speech and accent variation across dialects \citep[e.g.,][]{Bartelds2022,Yang2023}. However, the caveats of these deep learning-based features are that they integrate information across time spans and thus mix information from linguistic units of very different grain sizes. As a consequence, the feature vector for a single spoken word token also typically encodes information about surrounding words or phrases. Such a mix of information can be advantageous for optimizing task performance (e.g., distinguishing accents), but makes interpretation harder when studying form–meaning relations at the word token level. 

We therefore opted for constructing speech-derived form vectors directly using MEL spectrograms. At this point, the problem arises that speech tokens in spontaneous speech vary in duration. Creating spectrograms from words with different durations will lead to spectrograms with very different sizes, and will mask similarities in the shapes of spectrograms.

One possible solution is to derive (continuous) frequency band summaries and zero-pad them into fixed-length vectors \citep{Arnold2017,ShafaeiBajestan2021}. However, instead of zero-padding, when constructing a MEL spectrogram, we dynamically adjust the step size, a parameter determining how far the analysis window advances before the spectral values of the next frame are calculated. Across word tokens of different lengths, we adjust the step size such that spectrograms with a fixed number of timesteps are obtained.  Similarities in the shape of spectrograms emerge straightforwardly in such \textit{time-normalized spectrograms}. These time-normalized spectrograms incorporate information about word duration. In section~\ref{sec:word-duration}, we show that when duration is taken out of the time-normalized spectrograms using Partial Least Squares, results remain basically the same. 

\subsubsection{Constructing time-normalized spectrograms}

Time-normalized spectrograms are constructed as follows. For a given audio token, we (1) determine an optimal step size producing 50 time steps, (2) transform the speech signal into a spectrogram with the found step size, (3) convert the spectral information of each time step into 21 mel-frequency banks, and (4) vectorize the time-normalized spectrogram into a 1,050-dimensional speech vector. 

\begin{figure}[p]
\centering
\includegraphics[width=0.9\textwidth]{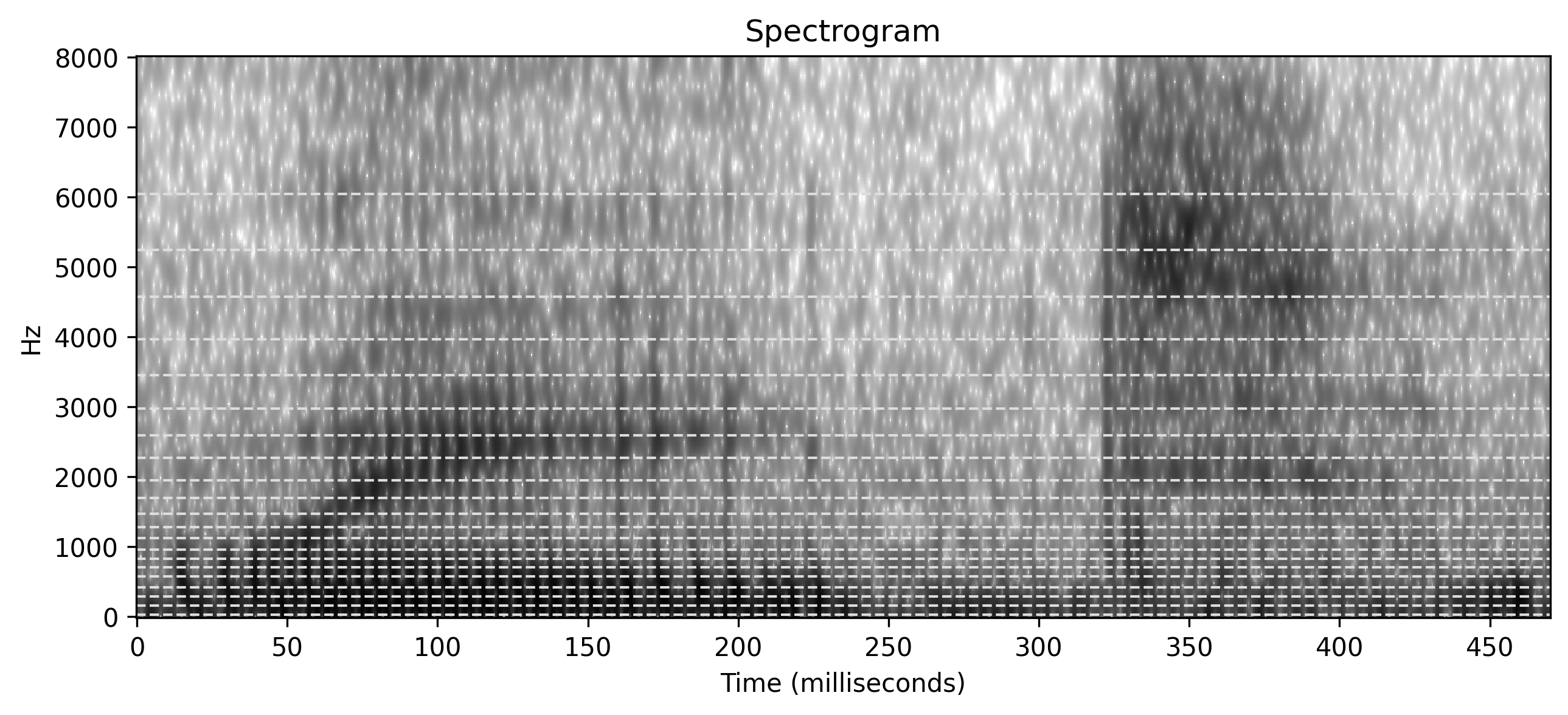}

\includegraphics[width=0.9\textwidth]{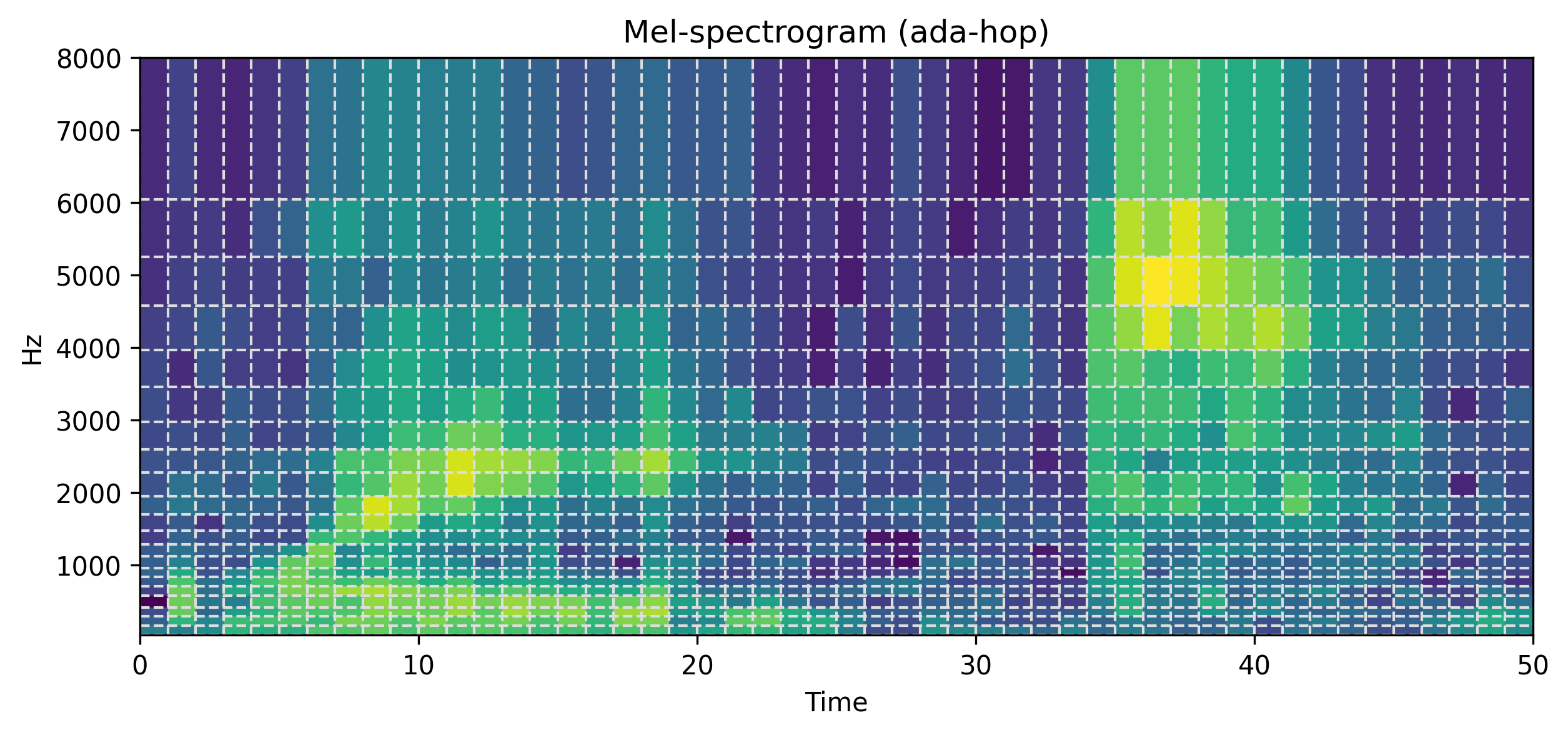}
\caption{
Spectrogram of a 470ms ``wait'' token (upper panel) and the corresponding fixed-size spectrogram with adaptive hop length (lower panel). The horizontal dashed lines indicate the frequency steps for the 21 Mel-frequency banks. The vertical dashed lines represent the 50 timesteps that are obtained with a step size of 9.4ms. 
}
\label{fig_ada_spec_new}
\end{figure}

Processing begins with the speech signals, which are time-domain digital waveforms representing physically continuous audio streams. From each speech signal, we derive a standard spectrogram (see the upper panel of Figure \ref{fig_ada_spec_new} for an example), a matrix that represents both temporal information by column (along the x-axis when visualized as a rectangular bitmap) and spectral information by row (along the y-axis). Each column represents the spectrum computed on a stretch (window) of the speech signal. The offset between consecutive windows, the step size, determines the number of columns in a spectrogram. If the step size is small, the window moves slowly along the speech signal and produces many columns; by contrast, if the step size is large, the window strides quickly and the spectrogram has fewer columns. We control the number of columns in a time-normalized spectrogram by determining an optimal step size, using the following equation:
\[
\text{step size} = \left\lfloor \frac{N - w}{T - 1} \right\rfloor
\]
The operator $\lfloor x \rfloor$ indicates the floor function, finding the largest integer smaller than $x$. $N$ is the number of samples in the speech signal, $T$ refers to the target number of frames, which is set to 50, and $w$ is the window size set to 10ms (i.e., 160 samples under a sampling rate of 16,000 Hz). Concretely, for a token with 7,520 samples (470ms), the equation gives us a step size of $(7520 - 160) / (50-1) \approx 150$ samples (9.4ms). Using the found step size, we convert the speech signal into a spectrogram of 50 columns (the temporal axis). On the spectral axis side, the number of rows is determined by the window size, which is set to 10 ms and yields 80 rows at a sampling rate of 16,000 Hz.\footnote{A 10 ms window corresponds to 160 samples at a 16 kHz sampling rate; after the Fourier transform, only one half of the symmetric spectrum is retained, yielding 80 rows.} However, the frequency sensitivity of the human cochlea varies across the frequency range. By compressing 80 rows into 21 rows using MEL-frequency filters, we derive a more compact spectrogram that better reflects the frequency sensitivity of the human cochlea \citep{Greenwood1990}. After these steps, the speech signal is now transformed into a fixed-size MEL-spectrogram: 50 columns in the temporal axis, and 21 rows in the spectral axis (see the bottom panel of Figure \ref{fig_ada_spec_new}).  Finally, we flatten the spectrogram by reshaping it into a 1,050-dimensional speech vector through row-wise concatenation. \footnote{These signal processing steps are carried out in Python with librosa \citep{McFee2015}. We also applied pre-emphasis (with a coefficient of .97) as a preprocessing step, with an effect similar to setting pre-emphasis to 6 dB/oct in Praat's spectrogram display.}

\subsection{Linear mappings between form and meaning}
\label{sec:LDL}

The alignment between the speech vectors, in form space, and CEs, in the meaning space, is investigated in the present study using the linear mappings of the DLM \citep{baayen2019discriminative,Heitmeier2025}.  Let $n$ denote the number of word tokens under investigation. For comprehension, a linear mapping $\bm{F}$ transforms the $n \times 1050$  form matrix $\bm{C}$ into a corresponding $n \times 768$ semantic matrix $\bm{S}$.  Conversely, a production mapping $\bm{G}$ starts with the semantic matrix and transforms it into the form matrix. As the mappings are matrices and no intermediate representation or non-linear functions are involved, they can be readily estimated using linear algebra \footnotemark:

\footnotetext{As least-squares solutions, these mappings minimize fitted error under the L2-norm, which implicitly assumes that all dimensions are equally meaningful, i.e., they live in a Euclidean space. This approximation works well in practice for the present data. More elaborate methods could, in principle, account for the covariance structure, e.g., minimizing Mahalanobis distance, or move beyond Euclidean geometry by considering geodesic distance on non-Euclidean manifolds, but these formulations are beyond the scope of the present study.}

\begin{align*}
\bm{F} &= (\bm{C}^\top \bm{C})^{-1}\bm{C}^\top \bm{S} \\
\bm{G} &= (\bm{S}^\top \bm{S})^{-1}\bm{S}^\top \bm{C}.
\end{align*}
\noindent where $\bm{C}^\top$ indicates the transpose of $\bm{C}$ and $\bm{C}^{-1}$ is the inverse of $\bm{C}$. The learned $\bm{F}$ and $\bm{G}$ matrices are used to compute the predicted semantic matrix, $\hat{\bm{S}}$, and the predicted form matrix, $\hat{\bm{C}}$, respectively:

\begin{align*}
\bm{C} \cdot \bm{F} &= \hat{\bm{S}}\\
\bm{S} \cdot \bm{G} &= \hat{\bm{C}}.
\end{align*}

The quality of these Linear Discriminative Learning (LDL) mappings can be evaluated by measuring how close the $\hat{\bm{S}}$ or $\hat{\bm{C}}$ matrices are to the actual $\bm{S}$ or $\bm{C}$ matrices. One such measure is by-word nearest neighbor accuracy \citep{Heitmeier2025}. For instance, when the comprehension mapping places a token's speech vector in the semantic space as a predicted or estimated CE, then, if the nearest neighbor in the semantic space belongs to the same word type, we count it as a correct prediction. Similarly, for production, spoken word tokens are represented as speech vectors derived from time-normalized spectrograms in form space, and a prediction is counted as correct if the nearest neighbor of a predicted speech vector belongs to the same word type. If the proportion of such correct predictions is higher than chance in the held-out data, this indicates the linear mapping has succeeded in capturing some of the similarity between the form and meaning spaces.

%
%

\section{Modeling the phonetic realization of homophones}\label{sec:modeling}

In this section, we carry out a series of experiments investigating whether homophone members can be distinguished by their phonetic realization, represented as speech vectors, and how speech vectors and CEs are aligned in form and meaning space. Section~\ref{sec:data} introduces the dataset that we investigated. 
Section \ref{speech-vec-lda} investigates whether the identity of a homophone pair can be predicted from the time-normalized spectrograms of its tokens, while Section \ref{speech-vec-lda2} asks whether the individual members within each pair can also be predicted. Section \ref{spectrogram-gam} makes use of the generalized additive model to clarify how the time-normalized spectrograms of homophones differ.  Next, in Section \ref{sec-alignment}, we implemented the LDL mappings between the speech vectors (flattened time-normalized spectrograms) and contextualized embeddings (CEs) to investigate the potential role of meanings in these word-specific realizations. We compared the alignments with permutation baselines (Section \ref{sec-exp-4.1} and \ref{sec-exp-4.2}), and show that the prototypical realization of a homophonic word can be constructed from the prototypical meanings of that word.

\subsection{Data}\label{sec:data}

The spoken word tokens are selected from the Redhen 2016 Dataset\citep{Joo2017}, which includes videos of United States television news broadcasts, corresponding subtitles, and the word-level timestamps obtained from a forced aligner \citep{ochshorn2017gentle}. Next, we identified the heterographic homophone pairs where the two word types met the following criteria: (1) each word type occurs at least 200 times; word types with more than 200 tokens were downsampled to 200 tokens, and (2) each token appears in a unique context, defined by up to 10 preceding words. Requiring each token to appear in a unique context removes artifacts caused by repeated advertisement phrases that occur across programs. These selection criteria led to a dataset with 35 homophone pairs, with 70 orthographic word types and 14,000 tokens. 

The 14,000-word tokens originated from 215 unique news programs. On average, a word type occurs in 63.2 programs. The mean duration of these tokens is 330ms ($SD = 117$ms). The distribution of spoken word durations has a slight positive skew: the first quartile ($Q1$) is 250ms, the median is 320ms, and the third quartile ($Q3$) is 400ms. The step size selected for time normalization of the tokens has a mean of 6.5ms ($SD=2.4$ms), with $Q1=4.9$ms, median$=6.3$ms, and $Q3=7.9$ms.

A minority of homophone pairs in this data set (8) are words with different morphological structure, such as \textit{knows} and \textit{nose}, \textit{band} and \textit{banned}, or \textit{passed} and \textit{past}. Three comprise irregular verb forms (\textit{seen, knew, blew}). However, most of the pairs of homophones (24) consist of monomorphemic words, such as \textit{wait} and \textit{weight}, \textit{mail} and \textit{male}. Some differences in spoken word duration may be expected due to differences in morphological structure and word category, as suggested by studies such as \citet{Seyfarth2017} and \citet{Lohmann2018}. Spoken word duration, however, is also co-determined by word meaning, as \citet{Gahl2024} showed using context-independent embeddings. In what follows, we first move beyond word duration to investigate systematic differences in how homophones are articulated, and how these differences covary with context-specific word meanings. The role of word duration will be revisited later, when we examine whether the observed differences remain after accounting for durations.







\subsection{Experiment 1: Predicting homophone pairs from speech vectors}
\label{speech-vec-lda}


Experiment 1 addresses the question of whether homophone pairs can be properly distinguished on the basis of their speech vectors.  This is a minimal requirement for the speech vectors that we constructed: if a homophone pair such as \textit{wait} and \textit{weight} cannot be distinguished from a homophone pair such as \textit{sell} and \textit{cell}, then the speech vectors that we constructed cannot be expected to be revealing about potential differences in the phonetic detail between the members within a homophone pair.

We used Linear Discriminant Analysis \citep[LDA, ][]{Fisher1936,Rao1948} with homophone pairs as a 35-level response factor, and the speech vectors as predictors. Given the correlational structure in spectrograms, we used Principal Components Analysis (PCA) to reduce the dimensionality of the speech vectors to 50, as the first 50 principal components account for 90\% of the variance in the original 1,050-dimensional speech vectors. These more compact speech vectors alleviate  the computational burden of handling high-dimensional vectors, while at the same time suppressing potential noise in the data \citep{Jolliffe2010}. 

We performed 10-fold cross-validation to assess how well homophone pairs can be predicted from the 50-dimensional speech vectors. Mean accuracy was 79.30\% ($SE = 0.42\%$). A permutation baseline, where we perform 10-fold cross-validation on a randomly permuted dataset, is 2.82\% ($SE = 0.10\%$). Most misclassifications involve pairs with highly similar pronunciations, such as \textit{seen} and \textit{sea}, \textit{wait} and \textit{way}, \textit{plane} and \textit{planes}. These results clearly show that the speech vectors that we constructed, perhaps unsurprisingly, contain rich information for discriminating between pronunciations.

\subsection{Experiment 2: Predicting homophones from speech vectors}
\label{speech-vec-lda2}

\begin{figure}[h]

  \centering

  \includegraphics[width=0.9\textwidth]{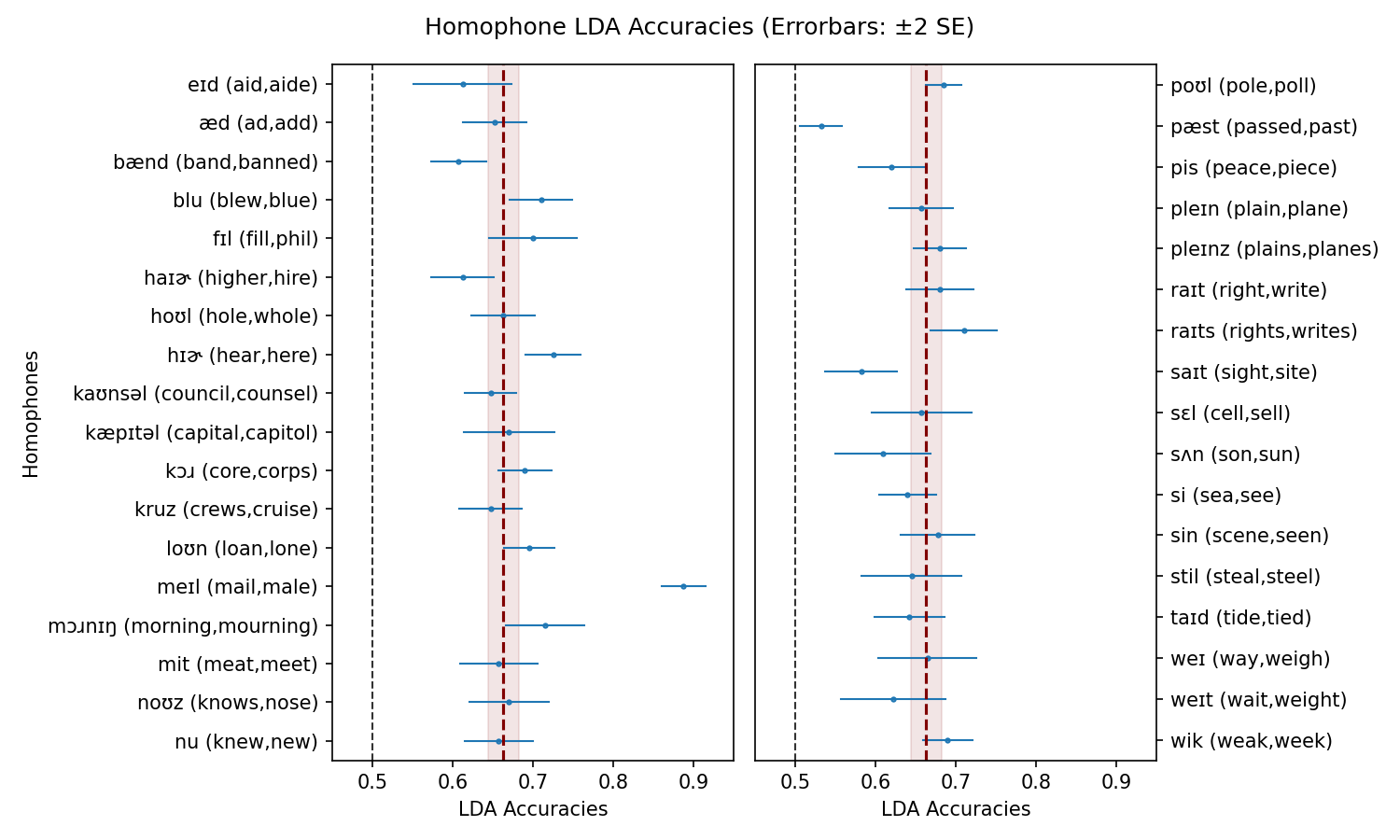}
    \caption{LDA classification accuracies for the speech tokens of the members of homophones within homophone pairs. The black vertical dashed line is the 50\% baseline, the red vertical dashed line represents the overall mean accuracy with a 95\% confidence interval in red, and the horizontal blue lines indicate 2 standard errors calculated with 10 cross-validation runs. Mean accuracies are represented by blue dots.}
  \label{fig_all_lda}
\end{figure}

Using the same procedure and method as for Experiment 1, we next ask whether a speech vector can distinguish homophone members across all pairs, using an LDA classifier. Each speech vector is classified into one of the 70 possible orthographic words. This is a much harder task, not only because the number of categories to be distinguished doubles, but also because if the members of a homophone pair are truly homophonous, the classifier should not be able to distinguish them (e.g., between \textit{wait} and \textit{weight}). Mean accuracy for 10-fold cross-validation was 51.13\% ($SE=0.46\%$). The corresponding permutation baseline was 1.43\% ($SE=0.11\%$). To rule out the possibility of random guessing within homophone pairs, we conducted additional LDA analyses, one for each homophone pair. Classification results are shown in Figure \ref{fig_all_lda}.  All homophone pairs are at least two standard errors above the 50\%-chance level, indicating spoken word tokens, despite being homophonous, can be classified into their word types significantly better than chance. These findings are consistent with previous studies documenting that heterographic homophones have, on average, different spoken word durations \citep{Gahl2024,Lohmann2018}. This experiment clarifies that the speech signals of members within homophone pairs differ not only in duration but also in their spectrograms: even a straightforward linear classifier can tell \textit{wait} and \textit{weight} apart with far above-chance accuracy.  Given the immense variability in speech, this is a surprising result.

This result cannot be simply explained by a potential difference between the members of the pairs in their utterance position. For each homophone pair, we computed proportion test statistics for tokens preceded by and followed by a pause, and correlated these statistics with LDA classification accuracies. Neither correlation was significant: $r=.26$, $p=.14$ for preceding pauses, and $r=.10$, $p=.58$ for following pauses. These results suggest that the classification accuracy cannot be simply attributed to differences in utterance positions.

While the LDA analyses show that differences in the pronunciation of homophones exist, they are uninformative about how members of a given homophone pair differ from each other. To address the question of how the spectrograms of homophones differ, we made use of the Generalized Additive Model \citep[GAM, ][]{Wood:2017}. 

\subsection{Experiment 3: Difference spectrograms}
\label{spectrogram-gam}

In order to clarify in what way the spectrograms of the members of the homophone pairs differ, we made use of the Generalized Additive Model \citep[GAM, ][]{Wood:2017}. In general, GAMs make it possible to statistically evaluate whether the estimated spectrogram surfaces for the two members in a homophone pair differ. For the present data, for any given homophone pair, the question is whether there is sufficient evidence to argue that the spectrogram predicted on the basis of the tokens of the first homophone (e.g., \textit{weight}) is significantly different from the spectrogram predicted on the basis of the tokens of the second homophone (e.g., \textit{wait}).  

The dataset we constructed for the GAM takes, for each homophone, its 1,050-dimensional speech vector as 1,050 observations, each of which is tied to a timestep (position on the temporal axis) and a frequency band (position on the spectral axis). For each observation, information was added about the duration of the token, the presence or absence of a preceding or following pause, and the speaking rate. For example, the datasets for \textit{weight} and \textit{wait} were combined into one dataframe with 420,000 rows and 7 columns \footnote{We use R and mgcv package for the Generalized Additive Model analysis}.

The analysis that we implemented sets up a smooth regression surface (predicted spectrogram) for one homophone, in combination with a difference surface (the predicted difference spectrogram) for the second homophone. When the difference spectrogram is added to the spectrogram of the first homophone, the predicted spectrogram of the second homophone is obtained. We modeled the response variable, the spectrogram value \texttt{spec}, as a smooth interaction of \texttt{timestep} and \texttt{filter\_bank}, using a tensor-product smooth (\texttt{te}). 

\begin{eqnarray}
\text{spec} & \sim & \text{te(timestep, filter bank)} + \nonumber \\ 
       & & \text{te(timestep, filter bank, by=I)} + \nonumber \\
       & & \text{s(duration)} + \text{s(speaking rate)} + \nonumber \\
       & & \text{preceding pause} + \text{subsequent pause} \label{eq:GAM1}
\end{eqnarray}

\noindent
As can be seen in this model specification, two tensor product smooths are requested, one for the first homophone, and a second one for the difference surface between the second homophone and the first. The variable \texttt{I} in the second smooth is an indicator variable that is 0 for the first homophone, and 1 for the second homophone. If a difference surface is statistically supported (in the present study, by a low empirical Bayes $p$-value), then the difference surface can be used to clarify which regions in the spectrogram are significantly different between the homophones of a homophone pair.  The above model also includes the abovementioned covariates (spoken word duration, speaking rate) and factorial controls for the presence of a preceding or following pause. 

%

\begin{figure}[htbp]
  \caption{An example of a GAM for a homophone pair: \textit{wait} vs. \textit{weight}. (a) The averaged spectrogram of \textit{weight}; (b) the averaged spectrogram of \textit{wait}; (c) the phone probabilities of \textit{wait/weight} in normalized time; (d) the difference spectrogram.}
  \centering
  \label{fig_wait_weight}
  \includegraphics[width=0.9\textwidth]{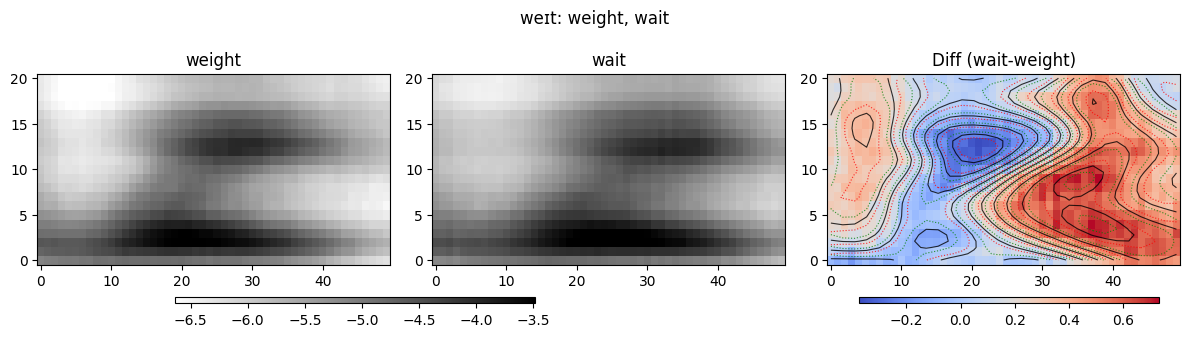}
\end{figure}

Figure~\ref{fig_wait_weight} illustrates key results for the homophone pair \textit{wait} and \textit{weight}. The two panels on the left present the mean spectrograms of the two homophones.  The two dark bands in the middle of these average spectrograms reflect the formants of the diphthong \textipa{eI}. The lower band around fbank 3 (centered at 416Hz) presents F1, and the upper band around fbank 12 (centered at 1952Hz) presents F2.  The coda \textipa{t} is not clearly observable in the mean spectrogram, suggesting it may not be fully articulated in the spontaneous speech under investigation here.  

Although, as expected, the two spectrograms are similar in many ways, some differences are already accessible to a visual scan.  These differences are brought out in a precise way by the difference spectrogram (i.e., difference surface) estimated by the GAM for \textit{wait} as compared to \textit{weight}.  Below, we return to the details of the color coding. For now, the black contour lines and their credible confidence regions (red dots: 1-SE down, green dots: 1-SE up) are of interest. If the 1-SE confidence contour regions of two neighboring contour lines overlap, it indicates that the corresponding contour lines are not reliably distinguishable, and hence the gradient between them is not statistically well supported. Here, however, we first note these confidence contour regions hardly overlap, from which we can conclude there are reliable gradients in the difference surface. Second, we see that during the first 10 timesteps, there is some more energy in the spectrogram of \textit{wait}, that around timestep 20, there is substantially less energy in the spectrum of \textit{wait}, and that around timestep 35, there is much more energy in the spectrogram of \textit{wait}. There is also a difference in timing of the vowel clearly visible in the difference surface: the energy of the nucleus is first reduced (shown in blue) and subsequently increased (shown in red), indicating a delayed onset of the vowel in \textit{wait} relative to \textit{weight}.


In order to assess whether the difference surface is statistically well-supported, we compared the GAM specified in equation~\ref{eq:GAM1} with a simpler GAM,
\begin{eqnarray}
\text{spec} & \sim & \text{I} + \text{te(timestep, filter bank)} + \nonumber \\ 
       & & \text{s(duration)} + \text{s(speaking rate)} + \nonumber \\
       & & \text{preceding pause} + \text{subsequent pause} \label{eq:GAM12}
\end{eqnarray}
in which both words have the same smooth surface for time by filter bank, but in which the indicator variable (I) allows for a difference in average intensity between the two words.  The difference in AIC of the two models was $\Delta$ AIC = 5208.73 in favor of model (\ref{eq:GAM1}),  which implies that the model that includes a difference surface is substantially more likely to have generated the data.


We carried out the same GAM analysis for all 35 pairs of homophones. The full set of 35 figures, showing averaged time-normalized spectrograms and their difference surfaces, is provided in Appendix A. Qualitative differences between homophone members are visible there; here we focus on quantitative differences. Across all 35 pairs, substantial reductions in AIC were present for the models that included a difference surface, indicating a strong word effect. The average AIC reduction of the \textit{word} models across homophone pairs was 3413.92 ($SD = 1881.87$). The homophone pair with the largest difference is \textit{mail/male} ($\dAIC = 8909.03$).  This pair also ranks among the highest in the LDA analysis (Section \ref{speech-vec-lda}). The homophone pair with the smallest yet still highly significant AIC difference is \textit{sight/site} ($\dAIC = 638.13$).  This pair also has low accuracy in the LDA classification.

These findings clearly indicate that these homophones have word-specific phonetic realizations, and LDA and GAM provide converging yet complementary evidence for this conclusion. The LDA demonstrates the contrast under a cross-validation perspective: this classifier generalizes from training to test data, which is possible only if there are systematic differences in the spectrograms of homophone pairs. The results obtained with LDAs are strengthened by the results obtained with the GAMs, which provide converging evidence from a statistical model-comparison perspective. Homophones must have word-specific realizations, as a model that includes a by-smooth for homophone outperforms a model that does not distinguish between homophones. In addition to identifying the spectral contrast surface, the GAM analysis also controlled for covariates such as duration and speech rate, showing that the word effect persists even when these covariates are included. Most importantly, the tensor-product smooth term provides a clear picture of how these words differ in terms of their mel-spectrograms.

Thus far, we have presented evidence that the spectrograms of homophones have homo\-phone\--specific properties. In the next section, we address the question of where these word-specific realizations originate from.


\subsection{Experiment 4: alignment between speech vectors and semantic vectors}
\label{sec-alignment}

Are the observed homophone-specific spectrograms aligned with their homophone-specific and utterance-specific meanings? Whereas in the preceding sections we represented word meaning categorically, by means of discrete class labels indicating homophone pairs or homophone members, we now turn to contextualized embeddings, which we expect to provide much more refined approximations of what words mean in their utterance contexts. Following the dimension reduction applied to the speech vectors, we also used PCA to reduce the dimensionality of the CEs to 50, with the first 50 principal components accounting for 96\% of the variance in the original 768-dimensional CEs.

%
%
%
%
%
%
%

\begin{figure}[p]
  \caption{Contextualized embeddings of homophone tokens. Each point stands for one word token's CE and is color-coded by word identity. All CEs are first transformed into word space using LDA and reduced into a two-dimensional space using t-SNE for visualization. Words are well-separated in the figure, consistent with the 95\% word classification accuracy based on CEs.   
  }
  \centering
  \label{fig_ce-tsne}
  \includegraphics[width=0.9\textwidth]{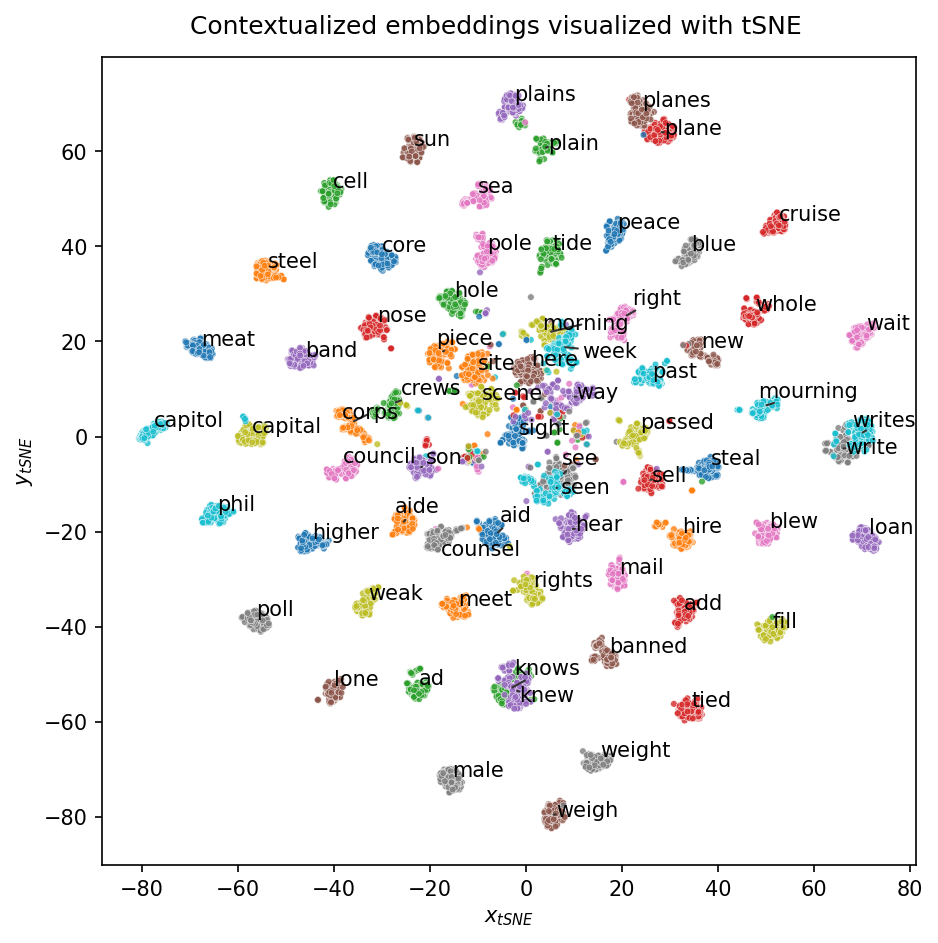}
\end{figure}

Figure \ref{fig_ce-tsne} visualizes the locations in the semantic space of the CEs of the 14,000 tokens of the 70 homophones in our dataset, using t-SNE \citep{maaten2008visualizing}. The by-word clusters are well-separated, which is consistent with the high word classification accuracy in LDA. In addition, the relative distances between words loosely reflect more general semantic similarities. For example, words in the same inflectional paradigm are close to each other, such as \textit{see} and \textit{seen}, \textit{knows} and \textit{knew}.\footnote{As the tokens used by GPT-2 for these words are disjunct, these similarities cannot be explained away as an orthographic artifact.}  Other pairs also show interesting proximities in the t-SNE map, such as \textit{counsel} and \textit{aid}, both relating to helping people; and \textit{sea} and \textit{tide}, both related to the ocean.  Furthermore, there is a clear tendency for nouns to have more negative values on the first t-SNE dimension, and for verbs to have more positive values on this dimension. 

Having clarified that the CEs indeed capture important aspects of meaning, we next discuss three computational experiments using these CEs.

\subsubsection{Experiment 4.1: LDL with full permutation}\label{sec-exp-4.1}

\noindent
We first investigated, for each homophone pair separately, the alignment between the speech vectors, in the form space, and the CEs, in the meaning space, using Linear Discriminative Learning (introduced above in section~\ref{sec:LDL}). We computed and evaluated both comprehension and production mappings for each homophone pair.

\begin{figure}[h]
  \caption{Alignments of predicted meaning vectors (left panel) and form vectors (right panel) compared to a full permutation baseline (in red). The vertical dashed lines represent mean accuracies. Confidence intervals are based on 10-fold cross-validation.}
  \centering
  \label{fig_alignment_full}
  \includegraphics[width=0.9\textwidth]{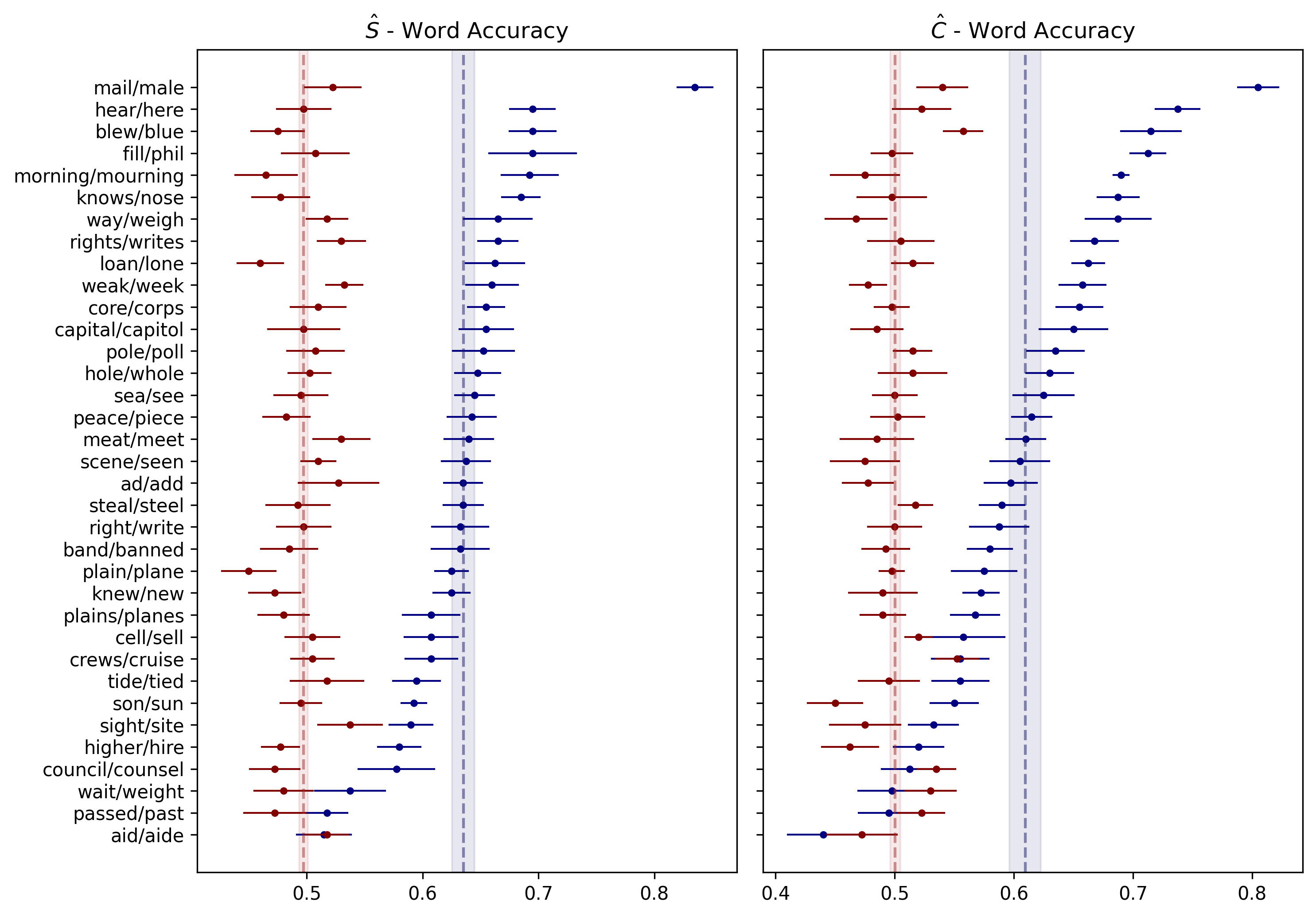}
\end{figure}

Each homophone pair has 400 tokens, comprising 400 rows in the respective form and semantic matrices. Each row in the form and semantic matrices is either a speech vector or a CE, both of which have 50 numbers. Thus, both the $\bm{C}$ and $\bm{S}$ matrices are of dimension $400 \times 50$. We used 10-fold cross-validation, evaluating a mapping $\bm{F}$ (or $\bm{G}$) with one fold of the data and learning from the remaining folds. As a baseline, we randomly permuted the rows of the 400 $\times$ 50 semantic matrix $\bm{S}$, and followed the same evaluation process.  Actual mapping performance can now be compared straightforwardly with a baseline in which the relation between form and meaning is broken. Results are presented in Figure \ref{fig_alignment_full}. Across comprehension (left panel) and production (right panel), LDL mappings perform significantly better for the actual pairing of form and meaning than the permutation baseline for most of the homophone pairs. Performance above the baseline indicates that the mappings, albeit simple linear ones, learn systematic regularities that are generalizable to unseen homophone tokens. The comprehension and production results both point to the same conclusion: form and meaning do not constitute two independent spaces. 


\subsubsection{Experiment 4.2: LDL with within-word permutation}
\label{sec-exp-4.2}

\noindent
In the preceding section, we observed that LDL accuracy is far above a full permutation baseline. What we do not know at this point is to what extent the tokens of a homophone are aligned with their own CEs.  Imagine two perfectly separable word clouds, each corresponding to one homophone of a pair of homophones (e.g., \textit{weight} and \textit{wait}).  If a mapping perfectly follows the token-level structure and places all predicted tokens exactly at their counterparts, classification accuracy at the word level will be perfect. However, if the mapping fully respects word structure without capturing any token-level correspondence, then each predicted token will still fall in the correct word cluster, also resulting in perfect accuracy.  Yet, in the second case, the mapping reflects only broader word-level structure, whereas the first provides stronger support for alignment between form and meaning at the individual token level.


For assessing the quality of form-meaning mappings between the spectrograms and CEs of homophones' individual word tokens, we proceeded as follows. First, we carried out within-word permutation, where the 200 semantic vectors of one homophone, e.g., \textit{weight} were randomly re-ordered, and those of the other homophone, \textit{wait} were randomly re-ordered separately.  As a consequence, the permuted semantic vectors of the tokens of \textit{weight} will always correspond to a speech token of \textit{weight}, albeit to a randomly selected token.

Second, we use a more informative index to evaluate the token-level alignment. The nearest neighbor word accuracy used above, defined as the proportion of predicted vectors whose nearest neighbor belongs to the same word type, is insensitive to the token-level structure, as it can be perfect even when the mapping captures only word-type distinctions. Therefore, we use the ratio $R_\text{L2}$, which assesses how much of the original vectors (speech vectors or CEs) is recovered by the LDL predictions. $R_\text{L2}$ is defined as the ratio between the sum of squared L2-norms of the predicted vectors $\hat{\bm{v}}$ and those of of the original vectors $\bm{v}$ across $n$ (here, $n=400$) observations: \footnote{
In the matrix case, e.g., $\hat{\bm{S}}$ or $\hat{\bm{C}}$, the sum of squared L2-norms of row-wise vectors $\bm{v}$ is identical to the squared Frobenius norm of the matrix. See \ref{appendix-Frobenius} for technical details.} 

\begin{equation}
R_\text{L2} = \frac{
    \sum_{i=1}^n \lVert \hat{v_i} \rVert^2
    }{
    \sum_{i=1}^n \lVert v_i \rVert^2
    }
  = \frac{
    \sum_{i=1}^{n} \hat{\bm{v}}_i \cdot \hat{\bm{v}}_i
    }{
    \sum_{i=1}^{n} \bm{v}_i \cdot \bm{v}_i
    }.
\end{equation}
To assess whether the $R_\text{L2}$ scores from actual data are significantly higher than those of within-word permuted data, we repeated the permutation process thirty times and constructed a 95\% confidence interval. An $R_\text{L2}$ score above the upper bound of this interval provides evidence for the existence of form-meaning alignment at the token level.

\begin{figure}[h]
\centering
\includegraphics[width=0.9\textwidth]{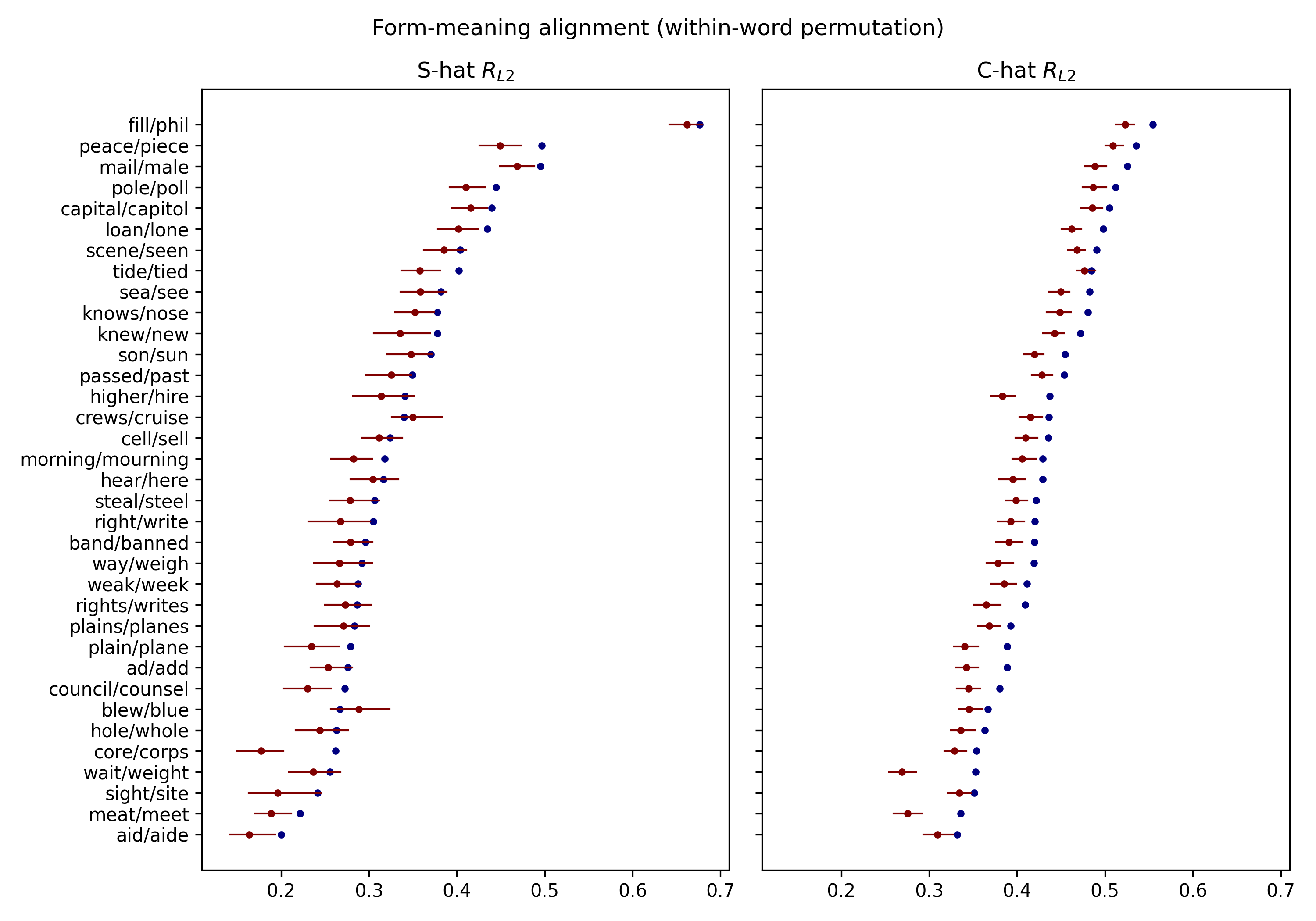}
\caption{$R_\text{L2}$ scores for comprehension (left panel) and production (right panel).  Blue dots represent empirical $R_\text{L2}$ scores, red dots and red horizontal lines represent 95\% credible intervals and means based on 30 within-word permutation cross-validation runs.
}
\label{fig-within-word-permute}
\end{figure}

Results are shown in Figure \ref{fig-within-word-permute} for comprehension (left panel) and production (right panel).  The blue dots represent the empirical $R_\text{L2}$ scores, the red dots and red horizontal lines represent the 95\% confidence intervals and means based on the 30 within-word permutation cross-validation runs. For production, almost all homophone pairs show above-chance form-meaning alignment, with only two exceptions. For comprehension, only 12 homophone pairs have observed values that lie outside the cross-validation interval. Nevertheless, there is only one pair where the empirical score is less than the cross-validation mean (\textit{blue}, \textit{blew}). 

Comparing the present findings with those reported in the previous section (\ref{sec-exp-4.1}), it is clear that token-level alignment is weaker than the type-level alignment.  This is unsurprising as the differences in the meanings of the homophones within homophone pairs are much larger than the more subtle  differences in the contextualized semantics of individual tokens of these homophones. 

The effects in comprehension are less pronounced, possibly because the comprehension models start from speech vectors that reflect individual speakers' own speech habits and emotions at the time of speaking. By contrast, the production models take CEs derived from a large language model that is trained on a vast amount of data and hence likely provides more stable and less variable representations.  As a consequence, the richer speech vectors contain information that is not matched in the semantic vectors, rendering prediction more difficult compared to the production model. With simpler `average' vectors as the starting point for production, the rich person-specific information in the speech vectors simply becomes just measurement noise.

The present results clarify that meaning variation below the level of word type plays a role in shaping phonetic realization. The subtle yet statistically significant token-level alignments shown in Figure~\ref{fig-within-word-permute} demonstrate that phonetic variation of homophones is not only conveying and realizing information about word type identities (e.g., \textit{weight} vs \textit{wait}), but also reflects above chance level subtle differences in the meanings of these word tokens as captured in the CE meaning space.

\subsubsection{Experiment 4.3: mapping from centroids}

\noindent We have shown that there is at least some alignment between the speech vectors of individual homophone tokens and their corresponding CEs. Next, we investigate whether the learned alignment gives rise to informative spectrotemporal structure by applying the learned production mapping to semantic centroids.

LDL learns the mappings through token-level semantic and speech vectors. Two possible outcomes may arise from these learned mappings: they either rely on subtle regularities that are generalizable without forming a clearly structured spectrotemporal pattern, or give rise to systematic patterns that, when contrasted between homophone members within a pair, are broadly consistent with the difference surfaces estimated from time-normalized spectrograms using GAM in Section~\ref{spectrogram-gam}. In this section, we examine the learned production mapping by applying the transformation matrix to prototypical semantic vectors, obtained by averaging semantic vectors within each homophone member, and ask to what extent the resulting predicted speech vectors correspond to the prototypical speech vectors.

Applying prototypical semantic vectors, i.e. \textit{centroids}, to the learned mapping matrix, e.g., $\bm{G}$ for the production model, follows naturally from a geometric point of view. The mapping matrix fully describes a schema with which the form space should be rotated, skewed, scaled, or optionally reflected, to best align with the semantic space. The prototypical vectors are valid inputs to the mapping matrix because they also live in the same semantic space on which the mapping matrix was estimated. Importantly, the $\bm{G}$ matrix works on both the observed and not-yet-observed vectors in the form space: the mapping is productive.



As a consequence, we can, for a given homophone (e.g., \textit{wait}) first calculate the centroid of its contextualized embeddings and then calculate its corresponding form vector predicted by the $\bm{G}$ mapping for that centroid.  The mapping has never seen this centroid during training, yet it can predict for the prototypical meaning of \textit{wait} what its spectrogram most likely looks like.   Similarly, we can calculate the centroid of the spectrograms of \textit{wait}, and map this prototypical spectrogram into the semantic space.  The expectation is that the semantic centroid and the form centroid align. 

If form and meaning centroids are indeed aligned \citep[for empirical evidence with respect to pitch contours in Mandarin Chinese, see][]{Chuang:Bell:Tseng:Baayen:2025,lu2025realization,Jin:Ernestus:Baayen:2026}, then the prediction follows that the empirical spectrogram difference surface of homophone pairs such as \textit{weight} and \textit{wait} detected by a GAM should be very similar to the difference between the predicted spectrograms of \textit{weight} and \textit{wait} obtained by applying the mapping $\bm{G}$ to the semantic centroids of these two words.

We tested this prediction as follows.  We first computed, for a given word type $w$, its corresponding semantic centroid $\bm{s}_w$ by averaging the CEs of all tokens of $w$. The resulting centroid represents the prototypical, context-independent, meaning of $w$.  We then applied the $\bm{G}$ mapping to $s_w$, which resulted in a predicted speech vector $\hat{\bm{c}}_w$.  As the mapping $\bm{G}$ is estimated using 50-dimensional speech vectors, in order to reconstruct the MEL spectrogram, we multiplied the 50-dimensional predicted speech vectors by the PCA basis vectors to obtain 1,050-dimensional vectors, which were subsequently reshaped into $50 \times 21$ matrices to recover the predicted MEL spectrograms. 

\begin{figure}[!htbp]
  \centering
    \begin{subfigure}[t]{.35\textwidth}
    \includegraphics[width=\textwidth]{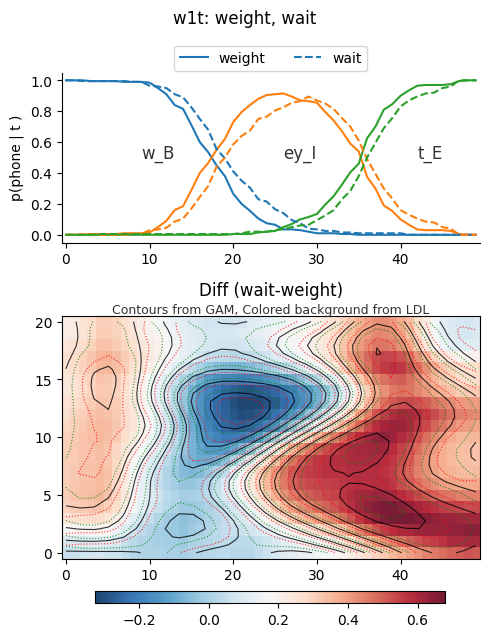}  
    \caption{\textit{wait} and \textit{weight}. }
    \label{fig_cpred_w1t}
  \end{subfigure}
  ~
  \begin{subfigure}[t]{.35\textwidth}
    \includegraphics[width=\textwidth]{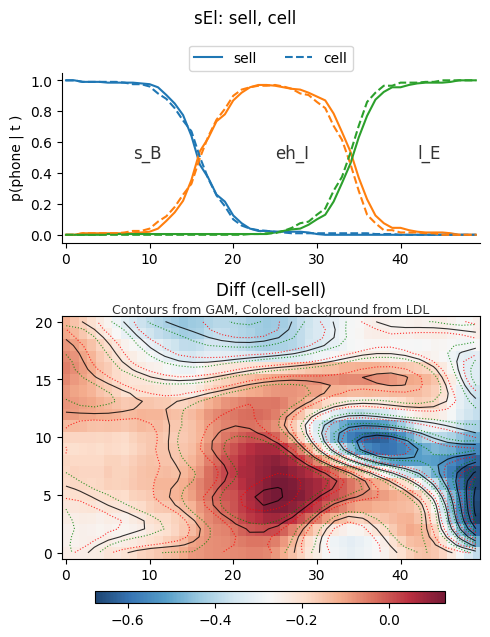}  
    \caption{\textit{sell} and \textit{cell}. }
    \label{fig_cpred_sEl}
  \end{subfigure}
  ~
  \begin{subfigure}[t]{.35\textwidth}
    \includegraphics[width=\textwidth]{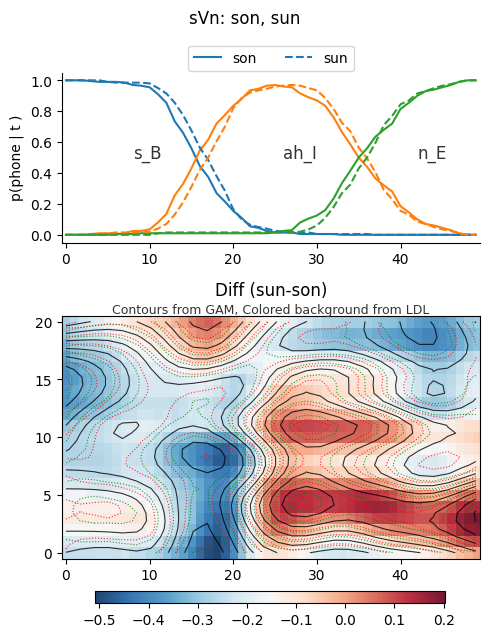}  
    \caption{\textit{son} and \textit{sun}.}
    \label{fig_cpred_sVn}
  \end{subfigure}
  ~
  \begin{subfigure}[t]{.35\textwidth}
    \includegraphics[width=\textwidth]{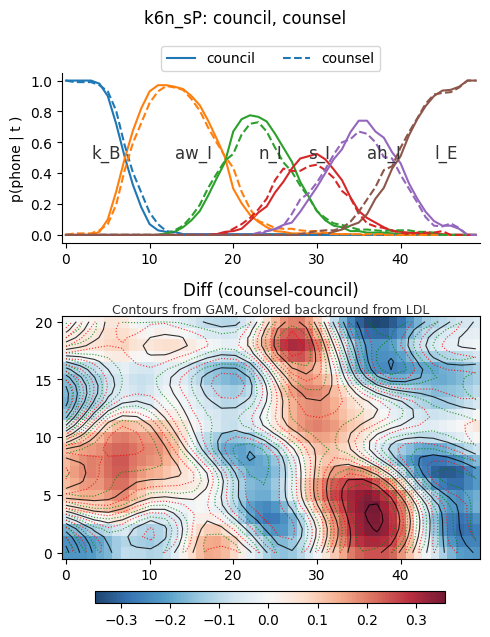}  
    \caption{\textit{council} and \textit{counsel}.}
    \label{fig_cpred_k6n_sP}
  \end{subfigure}
  
\caption{Selected difference spectrograms using GAMs applied to the observed spectra (contour lines) and using linear mappings from CE centroids (color coding). The close match between the predictions from LDL and the difference surfaces from GAM indicates that homophones' prototypical meanings co-determine their prototypical spectra. Upper panels show phone alignment profiles as references. Four homophone twins are shown: (a) \textit{sell} and \textit{cell}, (b) \textit{son} and \textit{sun}, and (c) \textit{council} and \textit{counsel}.}
\label{fig_pred_centroids}
\end{figure}

It turns out that contrast spectrograms obtained by taking the differences between two predicted MEL spectrograms generated from the semantic centroids are remarkably similar to the difference surfaces estimated with GAMs. Figure \ref{fig_pred_centroids} presents four homophone pairs as examples:  \textit{wait} and \textit{weight}, \textit{sell} and \textit{cell}, \textit{son} and \textit{sun}, and  \textit{council} and \textit{counsel}. The contour lines and their 1SE confidence regions are taken from the GAM models. The color coding reflects the differences in the spectrograms predicted from the homophones' CE centroids. The color coding and the GAM-based contour lines are well aligned. The correspondence is not perfect, as stripes in the colored background resulting from reconstruction errors are clearly visible. However, maxima and minima all align across figures, indicating that the difference surfaces estimated using empirical speech vectors alone can also be well approximated from the semantic space. 

The similarity between the contrast of predicted spectrogram centroids and the GAM-estimated difference surface is surprising, and shows that the linear mapping from embeddings to spectrograms is of high quality.\footnote{Given that a linear mapping exists between spaces A and B, linearity guarantees that the mapped centroid of A equals the centroid of the mapped points in B.} The contrast between the LDL-predicted spectrograms and the difference spectrograms estimated by the GAMs converges to a surprising extent. These results would not be possible without the presence of the structures within each homophone pair that were observed in Section \ref{sec-exp-4.1} and \ref{sec-exp-4.2}.


The examples shown in Figure~\ref{fig_pred_centroids} illustrate a further point, namely, that the differences in homophones' spectrograms are unlikely to reflect only differences in the timing of articulatory gestures within homophones. Figure \ref{fig_cpred_w1t} revisits the \textit{wait} and \textit{weight} pair discussed in Section \ref{spectrogram-gam}: the contrast spectrogram closely aligns with the GAM-estimated difference spectrogram, and the phone alignment profile, provided as a visual reference, consistently shows that listeners have to wait longer for the vowel onset in \textit{wait}. Additional examples are provided in Figure \ref{fig_pred_centroids}. Figure \ref{fig_cpred_sEl} displays the difference spectrogram for \textit{sell} and \textit{cell}, together with its phone alignment profile.  This phone alignment profile shows that there are hardly any differences in the timing of the segments of \textit{sell} and \textit{cell}.  Nevertheless, for this pair, there are substantial differences within the segments.  In Figure \ref{fig_cpred_sEl}, both the contrast spectrogram (colored background) and difference surface (contour lines) suggest more energy in the /s/ of \textit{cell} during the first five timesteps (darker red colors), after which this /s/ becomes softer (darker blue colors at frequencies above frequency band 18, which is around 4,576Hz).  The vowel of \textit{cell} is also articulated with more energy in the first and second formants, and the final liquid is overall realized with less energy, especially for lower frequency bands. These effects appear only in the spectral domain, as the durations of all three segments are well-matched in the temporal alignment profile.

A similar pattern is also observed for the \textit{sun}--\textit{son} pair (Figure \ref{fig_cpred_sVn}). Here, the /s/ of \textit{sun} shows not only a longer duration, but is also realized with stronger high-frequency energy, as indicated by the red area in the interval of 10--20 normalized time units. It is also clear that the first and second formants of the vowel carry more energy for \textit{sun}, and in the case of the first formant, it continues to do so for the final nasal. Within-segment variation can also be seen for \textit{counsel} and \textit{counsil} (Figure \ref{fig_cpred_k6n_sP}). For instance, the /s/ carries more fricative energy for \textit{counsel} than for \textit{counsil}, and the vowel of the second syllable of \textit{counsel} has more energy in the lower frequency bands.  

In summary, Figure~\ref{fig_pred_centroids} illustrates that the differences in phonetic realization of homophone pairs align with their contextualized embeddings. The learned alignment of embeddings and spectrograms in LDL captures informative structures, as shown by the structured contrasts obtained from semantic centroids, and by the similarity between these contrasts and the GAM-estimated difference surface derived from empirical speech vectors.

\section{Accounting for word duration}
\label{sec:word-duration}
Time-normalized spectrograms provide fixed-size representations for tokens that differ in spoken word duration. However, the duration may still leave traces in a time-normalized spectrogram, partly because vowels and consonants may be shortened differently in a spoken word \citep{Gay1978,Max1997,Janse2003}. This raises the question of whether the differences in the time-normalized spectrograms within homophone pairs are the consequence of spoken word duration itself, rather than the internal \textit{shape} of the spectrogram. To address this issue, we carried out two further analyses: one investigates the predictability of word duration from speech vectors derived from time-normalized spectrograms and semantic vectors; the other assesses whether the effects observed in earlier analyses still hold after removing duration from the speech vectors.

\begin{figure}[h]
\centering
\includegraphics[width=.9\textwidth]{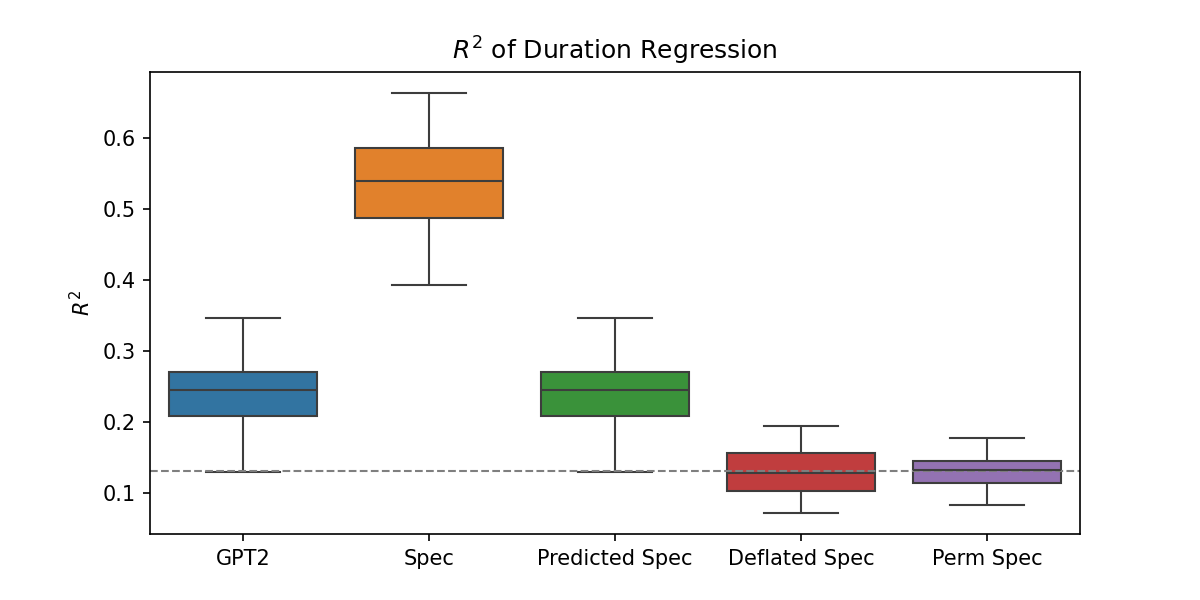}
\caption{\label{fig-dur-deflation} $R^2$ distributions across homophone pairs of regression models for duration. Predicting from embeddings (GPT2, blue) has a much lower $R^2$ compared to predicting from the speech vectors derived from time-normalized spectrum (Spec, orange). Unsurprisingly, predicting from the fitted speech vectors (Predicted Spec, green) has the same overall $R^2$ as predicting from embeddings. $R^2$ is lowest when predicting from deflated speech vectors (Deflated Spec, red) and similar to the $R^2$  obtained with permuted spectrograms (Perm Spec, brown).  Deflating the speech vectors successfully removes meaningful duration effects.
}

\end{figure}

Figure \ref{fig-dur-deflation} summarizes the results of the first analyses, which consider to what extent spoken word durations are predictable from spectrograms. This figure presents the distributions of $R^2$ values (across the 35 homophone pairs) obtained with 5 different regression models predicting spoken word duration. The mean $R^2$ for a regression model predicting duration from the GPT2 embeddings is 24.15\% (permutation baseline: 12\%), but when duration is instead predicted from the speech vectors (Spec, orange), $R^2$ increases to  52.90\%. This indicates that the speech vectors contain rich information about duration.  Unsurprisingly, the speech vectors predicted from the embeddings contain as much information about duration as the embeddings themselves, as shown by the green boxplot.

We used Partial Least Squares (PLS) to remove the information about duration from the speech vectors~\citep[][a brief introduction about PLS is also available in the appendix.]{Wold1966,Wold2001}. PLS iteratively identifies linear combinations of the predictors that are most strongly related to duration, and then removes these components, or \textit{deflates} them following the terminology in \citet{Wold2001}, so that the remaining predictors no longer contain enough information to predict duration beyond chance. The result of the PLS-deflated speech vector is shown as the red boxplot in Figure~\ref{fig-dur-deflation}, and its $R^2$ has dropped to 13.00\%, a value that is highly similar to that obtained with the permuted speech vectors that provide the baseline accuracy (12.74\%).


\begin{figure}[h]
\centering
\includegraphics[width=.9\textwidth]{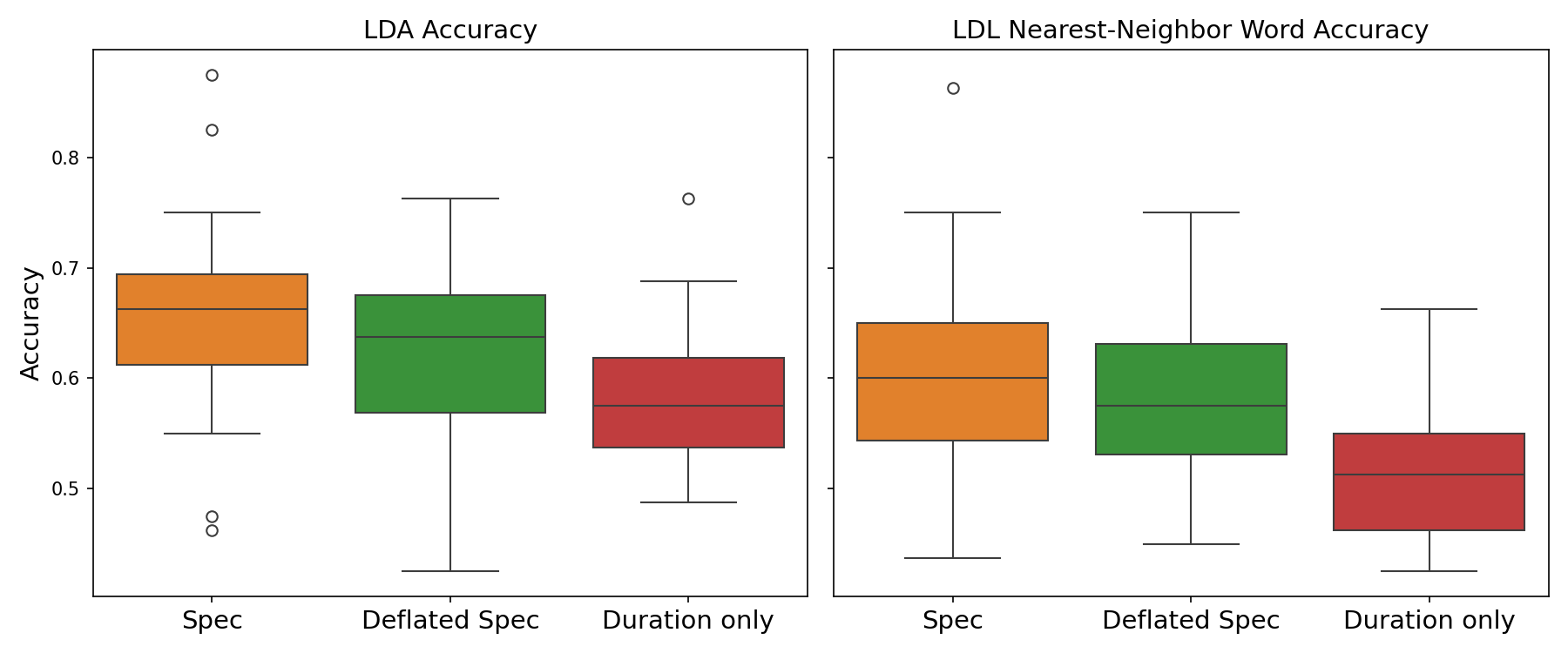}
\caption{\label{fig-dur-deflation2} Distributions (across the 35 homophone pairs) of LDA held-out accuracy (left panel) and nearest neighbor held-out accuracy (right panel) for models predicting from the observed spectrograms (orange), the deflated spectrograms (green), and from duration only (red). Removing duration from the spectrogram decreases prediction accuracy somewhat, but accuracy nevertheless remains high. 
}

\end{figure}

To assess the contribution of word duration in differentiating homophone members, the second set of analyses compares how well homophone members can be classified within homophone pairs (1) when the original time-normalized spectrogram is used, (2) when the classification is based on the PLS-deflated spectrogram from which information about duration is removed, and (3) when the LDA is given access only to word duration. Figure~\ref{fig-dur-deflation2} shows that predicting based on the PLS-deflated spectrogram decreases prediction accuracy ($M=62.50\%$, $SE=1.24$), an expected result given that duration is predictive for homophone discrimination \citep[see also][]{Gahl2008}. However, the decrease in accuracy is relatively small, and accuracy remains substantially higher than when homophones have to be classified on the basis of duration only ($M=58.1\%$, $SE=0.01\%$). Therefore, we can conclude that, consistent with the results shown in Section \ref{speech-vec-lda2}, the time-normalized spectrogram can still distinguish homophone members even when the word duration is controlled for.

Consistent results are also observed when mapping from speech vectors to semantic vectors using PLS-deflated speech vectors (Figure \ref{fig-dur-deflation2}, left panel). Compared to the original speech vectors from Section \ref{sec-exp-4.1}, the PLS-deflated spectrograms (orange) still align with the semantic vectors and can predict the semantic vector only slightly worse than the original speech vectors (green boxplot; $M=58.39\%$, $SE=1.23\%$). In the case of LDL mappings, the spoken word durations by themselves can only predict the semantic vectors at the chance level ($M=51.79\%$, $SE=1.12\%$). These results consistently show that word duration contributes to form-meaning alignment, but does not fully account for the alignment between speech and semantics.

In these analyses, we have considered whether the alignment between form and meaning can be traced back completely to spoken word duration. Our analyses do not support this argument, as time-normalized spectrograms from which duration has been removed remain highly predictive for the members of homophone pairs, indicating that the \textit{shape} of their spectrotemporal realization still plays an important role in form-meaning alignment.

\section{General Discussion}
\label{discussion}
\noindent
This study investigates the phonetic realization of English heterographic homophones and the role of word meaning in shaping the fine details of their pronunciation. To this end, we collected 200 tokens for 70 homophones of 35 homophone pairs from the Redhen 2016 dataset of US television news broadcasts. For each of the 14,000 tokens, we calculated the time-normalized spectrogram, together with a contextualized embedding using GPT-2.

Experiment 1 showed that the 35 homophone pairs can be teased apart with high accuracy (79.3\% vs. a 2.8\% permutation baseline) based on their vectorized time-normalized spectrograms. Experiment 2 clarified that LDA can discriminate with good accuracy (51.1\%, vs. a 1.4\% permutation baseline) between the 70 individual words (wait, weight, \ldots).  Furthermore, LDA analyses carried out for the 400 tokens of individual homophone pairs consistently predicted homophone identity (e.g., \textit{weight} vs. \textit{wait}) far above the 50\% baseline. This shows that there are systematic differences in the speech signals of homophones within homophone pairs.

In order to visualize in what way the spectrograms of homophones within a homophone pair differ, Experiment 3 made use of the generalized additive model (GAM), using tensor-product smooths. The 35 GAM models consistently pointed to substantial differences between the spectrograms of homophone twins (the average decrease in AIC obtained by including a difference spectrogram in the model specification was 3114). The GAMs highlight clear differences in the fine detail of phonetic realization between members in a homophone pair. For instance, for the homophone pair \textit{weight} and \textit{wait}, the listener has to wait longer for the vowel to arrive in \textit{wait} as compared to \textit{weight}. In another case,  for the \textit{sun} and \textit{son} pair, the /s/ in \textit{sun} is realized with stronger high-frequency energy than that in \textit{son}.
  
Experiment 4 addressed the hypothesis that the differences in the phonetic realization of members within a homophone pair can be traced back to their meanings using LDL. For each of the 35 homophone pairs, we constructed a comprehension mapping from spectrograms to CEs and a production mapping from CEs to spectrograms. These mappings were evaluated and examined in three sub-experiments. First, Experiment 4.1 evaluated the mappings using by-word nearest neighbor accuracy. Both mappings performed remarkably well, outperforming mappings trained on fully permuted embedding matrices by a wide margin. Next, in Experiment 4.2, we provided evidence supporting the alignment of form and meaning at the individual token level. For production, all homophone pairs yielded higher scores than the permutation baseline, and with only two exceptions, the empirical scores are well outside the 95\% permutation baseline confidence interval. This constitutes clear evidence for the CE space and the phonetic space being aligned to some extent even at the token level.

In Experiment 4.3, we showed that the learned alignment has an informative structure linking the most prototypical meaning to the most prototypical speech realization. By using the semantic centroid as input, the learned alignment predicted the corresponding spectrogram centroid, and the contrast between the two prototypical spectrograms closely resembled the difference surface estimated by GAMs. Taken together, the meanings of homophone members are aligned with their phonetic realizations both at the type level, where word identity can be predicted above chance and semantic centroids map onto spectrogram centroids, and at the token level, where variation in meaning is reflected in phonetic variation across individual tokens.

The potential confound of word duration was addressed in Section \ref{sec:word-duration}. We showed that, even after controlling for duration using partial least squares, the deflated representations, which no longer predict duration above chance level, still differentiate the members of a homophone pair and align with the semantic vectors, further supporting the results reported in Section \ref{speech-vec-lda2} and \ref{sec-alignment}. Importantly, these findings indicate that the realization differences observed in the speech representations cannot be fully attributed to differences in spoken word durations, the importance of which has been documented in previous studies \citep{Drager2011,Gahl2024}. Instead, they suggest that other fine-grained aspects of phonetic realization vary systematically with word meaning.

The present set of 35 homophone pairs is too small for teasing apart the effects of morphological structure \citep{Seyfarth2017} or word category \citep{Lohmann2018}.  In all likelihood, these effects of morphological structure and word category are part and parcel of the semantic effects that we have documented.  For instance, the differentiation of the contextualized embeddings of homophones in a t-SNE map (see Figure \ref{fig_ce-tsne}) shows some differentiation by word category, which suggests that our results are to some extent co-determined by the different semantics associated with nouns, verbs, and adjectives.

Two aspects of the present results are particularly surprising. First, contextualized embeddings provide only a rough approximation of the full richness of what individual speakers had in mind when uttering a homophone token, but they nevertheless align with the spoken word tokens. Second, the alignment can be described simply as a linear mapping, without requiring a more complex non-linear one. We will return to the linearity issue later in the discussion, but it already shows that the underlying isomorphism between form and meaning is significant.

More importantly, the differences in realization reported in this series of experiments are not merely descriptive patterns in the observed samples, but systematic regularities that generalize to unseen data. In other words, the point is not simply that the homophone members such as \textit{wait} and \textit{weight} differ, but that the time-normalized spectrograms contain systematic cues that make homophone members distinguishable. This regularity is also found in the alignment between form and meaning. The precise source of these regularities remains unclear, particularly given the richness of information carried by the speech signal. However, Experiment \ref{speech-vec-lda} shows that the effect cannot be attributed only to utterance position, and an exclusive effect of word duration is ruled out in Section \ref{sec:word-duration}. Although other factors may potentially influence speech, such as speaker identity, emotional states, stress, focus, prominence, and even force-aligner errors, they are unlikely to systematically contribute to a generalizable difference in 400 random samples per homophone pair, across 35 homophone pairs drawn from diverse contexts and broadcast sources. 

One possible factor underlying the generalizable difference of homophone members is the information content of different words. \citet{vanSon2003} reported that the higher the segmental information content of a segment, the less likely the segment is to be reduced. Segmental information content was operationalized by means of a corpus-based estimate of how distinctive a word is in context. Specifically, \citet{vanSon2003} computed the conditional probabilities of words occurring within the context surrounding the target word; the more distinctive the context of a target word, the lower its segmental information content and, in turn, the greater the tendency toward reduction, including slightly shorter duration.

Current results are compatible with this account: homophone members differ phonetically, possibly because each member's contextual distinctiveness differs. This difference can be observed throughout the speech signal (i.e., the time-normalized spectrogram) and is not limited to duration. Furthermore, the current study approaches the effect of context differently, namely, by using CEs, which can be viewed as a token-level semantic content representation computed from the preceding words rather than a scalar estimate of information content. These two notions are nevertheless related under the view of a large language model (LLM) as a compressor: information content corresponds to code length, whereas CEs can be understood as the internal predictive state from which such code is produced \citep[for literature of viewing LLM as a compressor, see][]{Deletang2024,Tseng2024}. Our analyses provide evidence that words' meaning in context, as captured by contextualized embeddings, predicts word tokens' spectrograms not only at the level of word types but also above chance level for individual tokens. In other words, both token-level measures of informativity and surprisal, and token-level contextual meaning capture aspects of how the cognitive system responds to speech as it unfolds.

It is possible that prosodic factors such as rhythm, loudness, intonation, prominence, and stress contribute to the segmental differences, but such an account is not very compelling, for two reasons. First, we found that the realization differences align with contextualized embeddings. Contextualized embeddings are based on text and cannot directly encode intonation, stress, rhythm, or loudness, unless these are already co-determined by the grammar and lexical choice inherent in the text. In that case, those features will be grounded in grammar and lexis \citep[for the reflection of grammar in contextualized embeddings,  see][]{Linzen2021,Manning2020}. The relation between contextualized embeddings and lexis is clearly visible in Figure~\ref{fig_ce-tsne}: the primary factor shaping the t-SNE map is homophone semantics. Those prosodic factors that are not themselves systematically grounded in grammar and lexis are unlikely to be captured by the contextualized embeddings, and hence cannot explain the observed isomorphy between the form and meaning spaces. 

Second, prosody can influence speech in a way independent from CEs, and we do not claim that CEs can completely reconstruct the spectrogram. Prosodic factors such as speech rate, the speaker's emotional state, and the mechanical constraints on articulation that underlie many sandhi processes are all contributing factors but are outside the scope of what can be captured by current CEs. What the present study provides is a detailed computational/statistical model for the extent to which articulation is co-determined by meaning in context. The residuals of this model are non-negligible, and thus present a challenge for follow-up research, namely, to provide equally explicit and equally falsifiable complementary predictions for these residuals that are derived from factors such as the speaker's emotional state, speech rate, register, focus, and the mechanical constraints on articulation.

We modeled form-meaning mapping with simple linear transformations, which appears to be all we need for predicting tokens' time-normalized spectrograms from the corresponding contextualized embeddings and vice versa; a similar conclusion with respect to the phonetic realization of tone in Mandarin Chinese was reached by \citet{Chuang:Bell:Tseng:Baayen:2025}.  Although the true form-meaning isomorphism underlying basic linguistic cognition \citep{huettig2025enhanced} may involve both linear and non-linear components, the success of linear transformations in the current study suggests that, at least within the local form and meaning space examined here, the contribution of non-linear components is limited. The relatively strong linearity that we observed in the present study may also result in part from the speech representations used, which are derived from mel-spectrograms and thus reflect some of the non-linear transformation performed by the human cochlea. The resulting outputs from the cochlea thus offer to the cognitive system representations that can be efficiently handled by linear mappings, which have broader cognitive advantages. They are simple and energy-efficient,\footnote{We thank Sebastian Otte for helpful discussion on this point.} and the learning rules for linear mappings are well-established across plant cognition \citep{gagliano2016learning}, animal cognition \citep{Rescorla:1982,Rescorla:1988}, and human language learning \citep{ellis2006selective,Heitmeier2023,Heitmeier2025}. Thus, linear mappings may offer an evolutionary advantage for cognitive processes for which efficient routinized execution is essential, as they are less computationally intensive and require less energy. 

The alignment of phonetic form and distributed meaning at the token level challenges a foundational assumption in linguistic phonology and psychology. First consider the classical models of \citet{Dell1986} and \citet{Levelt1999}. The finding that homophone twins have different pronunciations could perhaps be accommodated by assigning them different form nodes, although further ancillary measures would be required to ensure that their segments are realized with appropriate homophone-specific detail. However, the alignment of form and meaning at the token level argues against abstract word form units blocking meaning-in-context shaping segmental phonetic realization. Likewise, models of auditory word recognition  \citep[e.g.,][]{McQueen:Cutler:Norris:2006,Norris:McQueen:2008,Bosch2022} in which abstract phones play a pivotal role, are severely challenged by the isomorphy that we have shown to exist between the phonetic space and the semantic embedding space.

In their criticism of formal phonology, \citet{Port:Leary:2005} stated that ``Linguistics has mistakenly presumed that speech can always be spelled with letter-like tokens.'' (p. 927)  They argued against the common assumption that phonetic segments are formal symbols, and their study provides ample evidence for their case.  Our results likewise argue against abstract phones, not only in psychological models of lexical processing, but also in linguistic theory.  In the present study, we have used the term ``segment'' in a pre-theoretical and descriptive sense.  The fact that the spectrograms of homophone tokens can be predicted to a considerable extent from their contextualized embeddings provides a clear indication that the theoretical construct of the phone may be redundant.  Likewise, the fact that pitch contours on Mandarin words are aligned to a considerable extent with contextualized embeddings \citep{Chuang:Bell:Tseng:Baayen:2025,Jin:Ernestus:Baayen:2026,lu2025realization} obviates the need for discrete tonal units.  Furthermore, spoken word duration is also co-determined by semantics \citep{Gahl2024,Jin:Ernestus:Baayen:2026}. 

However, for modeling speech production, mappings from embeddings to time-normalized spectrograms, time-normalized pitch contours, or spoken word durations are no more (but also no less) than a promise of a proof of concept. Given that homophone duration can also be predicted from word embeddings as shown in Section \ref{sec:word-duration}~\citep[in combination with other factors such as speech rate and contextual probabilities, see][]{Gahl2024}, a word token's spectrogram in real time can in principle be predicted by combining mappings from embeddings to spectrogram shape and mappings from embeddings to duration. A more realistic production model can directly model articulators by predicting sets of time series of control parameters for the vocal tract that jointly give rise to the speech signal, instead of predicting the speech signal itself. Whether this is possible with simple linear mappings currently is an open question. Considerable headway in this general direction has been made with a deep learning architecture trained on spectrograms and embeddings in conjunction with a physical model of the vocal tract \citep{Sering:2023,Sering2024}.

The distinction between Basic Language Cognition (BLC) on the one hand and Extended Language Cognition (ELC)  \citet{hulstijn2024predictions} (renamed as the Enhanced Literary Mind (ELM) by \citet{huettig2025enhanced}) is useful for understanding the present results in a broader context. Hulstijn defines BLC as ``a person's ability to comprehend and produce spoken language in situations of everyday life, common to all adult native speakers in a given language community'' (p. 3). BLC is viewed as implicit, unconscious, experience-driven cognition, acquired through massive exposure that enables fast and automatic speech processing. The Discriminative Lexicon Model \citep{Heitmeier2025} was conceptualized as a computational proposal for basic \textit{lexical} cognition.  The DLM's production and comprehension mappings are subliminal, not open to introspection, and represent automatized processes that are continuously updated with experience \citep[cf.][]{Heitmeier2023}.  As a consequence, we predict that in meta-linguistic tasks, language users cannot tell homophones apart from just their audio signals. 

It remains an open question to what extent basic lexical cognition is affected by the cognitive processes that build on the knowledge and skills that come with training in literacy.  We cannot rule out that the heterographic homophones are articulated differently precisely because they have different spellings.  If the tokens of homographic homophones (e.g., \textit{bank} and \textit{bank}) are found to have discriminable contextualized embeddings but indistinguishable pronunciations, this would provide evidence against the Discriminative Lexicon Model.  

In summary, heterographic homophones are only approximately homophonous and differ not only with respect to their spoken word duration, but also with respect to a range of other aspects of their phonetic realization. The exact realization of a homophone token in normalized time is co-determined by its meaning in context, and can be approximated by a linear mapping from its contextualized embedding. Time-normalized spectrograms are a promising tool for probing the fine details of phonetic realization and obviate the need for phonetic transcriptions. The abstract sound symbols of formal phonology hide the truth of the remarkable alignment of phonetics with semantics from our eyes.



\appendix
\renewcommand{\thesection}{Appendix \Alph{section}}
\newcommand{\cdash}{\mbox{\rule[0.5ex]{0.8cm}{0.4pt}}}
\setcounter{section}{1}

\renewcommand{\thefigure}{A\arabic{figure}}
\setcounter{figure}{0}

\includepdf[pages=-, pagecommand={}]{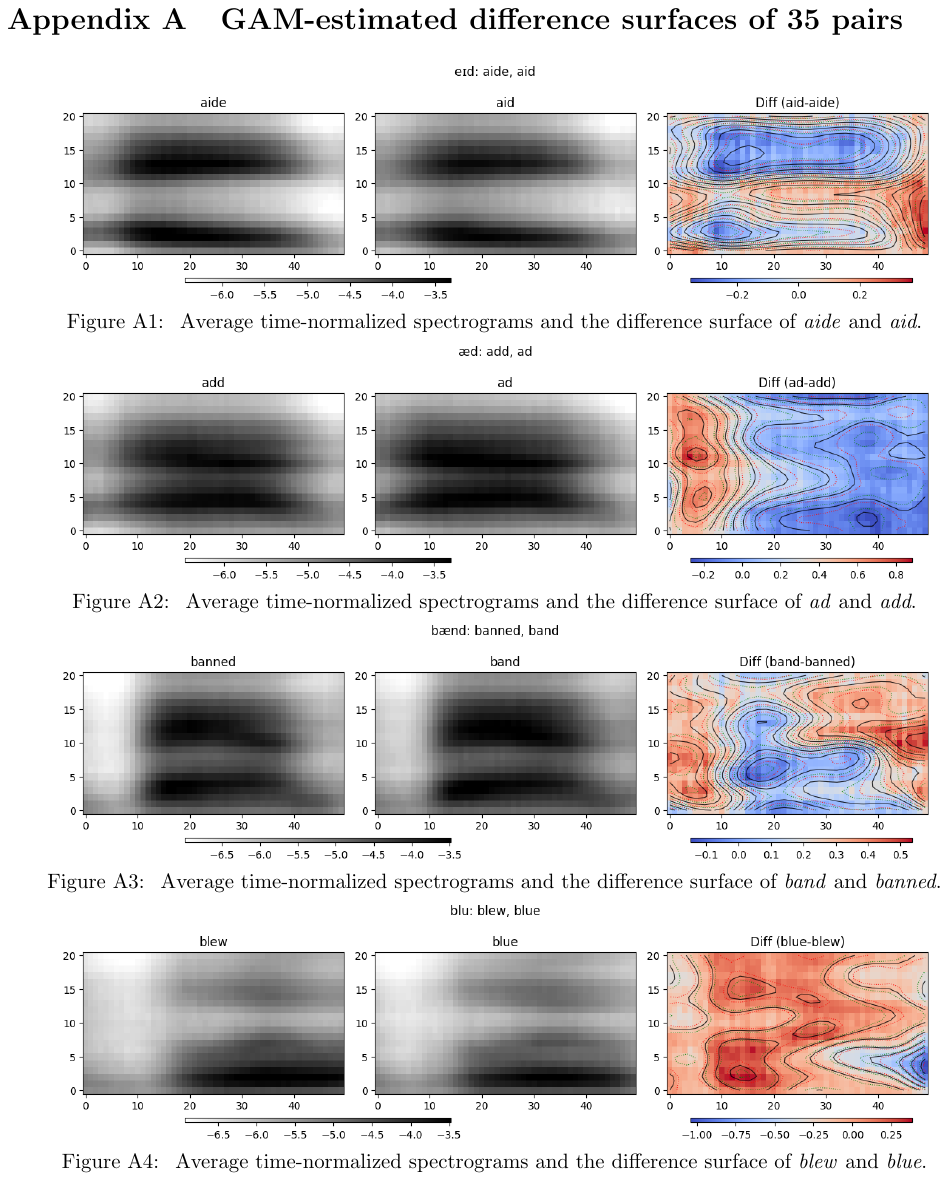}

\section{Partial Least Squares}

Partial least squares (PLS) is an iterative process. We write the original cue matrix $C$ as $C_1$, where the subscript indicates the first iteration. First, PLS finds a direction $w_1$ that maximizes the covariance between the spectrogram features $C$ and the durations. The score $t_1$ is the linear combination of the features with the found weights, $w_1$:  

$$
t_1 = C_1 w_1
$$

After finding the $t_1$, we can find the directions, or loadings, $p_1$, so that the corresponding rank-1 matrix, $t_1p_1^\top$, best approximates $C$:

$$
p_1^\top = \frac{t_1^\top C_1}{t_1^\top t_1}
$$

Finally, we can remove the rank-1 matrix from $C$. The scores that maximally covary with the durations are thus removed from the $C$.

$$
C_2 = C_1 - t_1 p_1^\top
$$

We repeat this process for 3 times, and obtain three weights, $w_1$, $w_2$, and $w_3$, and three scores, $t_1$, $t_2$, and $t_3$, along with the loadings, $p_1$, $p_2$ and $p_3$. Using these components, we can construct $C_1$, $C_2$, $C_3$ and $C_4$:

\begin{align*}
C_4 &= C_3 - t_3 p_3^\top \\ 
    &= C_2 - t_2 p_2^\top - t_3 p_3^\top\\    
    &= C_1 - t_1 p_1^\top - t_2 p_2^\top - t_3 p_3^\top \\
&= C_1 - 
\begin{bmatrix}
t_1 & t_2 & t_3 
\end{bmatrix} \begin{bmatrix}
p_1^\top \\
p_2^\top \\
p_3^\top
\end{bmatrix} \\
& = C_1 - TP^\top 
\end{align*}

One important property of the scores is that they are orthogonal to the
deflated matrices obtained in later iterations. To see that, we can use $t_1$ and $C_2$ as an example:

\begin{align*}
t_1^\top C_2 &= t_1^\top (C_1 - t_1 p_1^\top) \\
             &= t_1^\top C_1 - t_1^\top t_1(\frac{t_1^\top C}{t_1^\top t_1}) \\
             &= 0
\end{align*}

Using this result, we can further show that $t_1$ is also orthogonal to $C_3$,

\[
t_1^\top C_3
= t_1^\top (C_2 - t_2 p_2^\top)
= t_1^\top C_2 - t_1^\top (C_2 w_2) p_2^\top
= 0
\]

Following the same reasoning, all extracted scores are orthogonal to the final
deflated matrix $C_4$, that is,

$$
T^\top C_4 = \begin{bmatrix}
\cdash & t_1^\top & \cdash \\
\cdash & t_2^\top & \cdash \\
\cdash & t_3^\top & \cdash \\
\end{bmatrix} C_4 = 0
$$

We denote $C_4$ as $C_{\textrm{defl}}$, as this is the deflated speech representations which can no longer predict durations beyond the permutation baseline. Using the orthogonality derived above, the Frobenius norm of the original matrix $C_1$ can be decomposed into the sum of the Frobenius norms of $C_{\textrm{defl}}$ and the components extracted by PLS, $TP^\top$:

\begin{align*}
\lVert C_1 \rVert_F^2
&=
\lVert TP^\top + C_{\textrm{defl}} \rVert_F^2 \\
&=
\lVert TP^\top \rVert_F^2
+
\lVert C_{\textrm{defl}} \rVert_F^2
+
2\,\mathrm{tr}\left((TP^\top)^\top C_{\textrm{defl}}\right)
\\
&=
\lVert TP^\top \rVert_F^2
+
\lVert C_{\textrm{defl}} \rVert_F^2
\end{align*}

The cross-product term is zero because we have shown that $T^\top C_{\textrm{defl}} = 0$, and a rearrangement of the cross-product term becomes

$$
(TP^\top)^\top C_{\textrm{defl}} = P (T^\top C_{\textrm{defl}}) = 0
$$

This derivation shows that although PLS does not provide orthogonal bases, as the loadings $P$ are generally not orthogonal, the fitted component $TP^\top$ is nevertheless orthogonal to the deflated matrix $C_{\textrm{defl}}$. This still allows us to think of the decomposition in linear terms. For more background on the Frobenius norm, see \ref{appendix-Frobenius}.

Using this decomposition property, we can further quantify the role of the duration component in the two components found in the LDL production model: the prediction $\hat{C}$ and the residual $E$. The proportion of variance, measured by the Frobenius norm, accounted for by $\hat{C}$ is .43, and that of $E$ is .56. Consistent with Figure \ref{fig-dur-deflation}, the $R^2$ of duration regression for $\hat{C}$ is .24, and that for $E$ is .38. We can further decompose $\hat{C}$ using PLS into a duration component $\hat{C}_{dur}$ and $\hat{C}_{other}$. The proportion of variance of $\hat{C}_{dur}$ is .08, and that of $\hat{C}_{other}$ is .35, already showing that duration is not a major component in the form-meaning alignment. Most of $\hat{C}$, therefore, reflects information beyond duration. Looking further at the $R^2$ of duration regression clarifies where the duration effect comes from: using $\hat{C}_{dur}$ as predictors, the $R^2$ is .16, while that for $\hat{C}_{other}$ is .08. Therefore, part of the duration component is reflected in the semantic embeddings and is also mapped into the predicted spectrograms. This is consistent with the fact that duration is codetermined by semantics. However, a substantial part of the information unrelated to duration also contributes to the alignment.

Not all duration information can be captured by semantics, at least not by the contextualized embeddings derived from text. We again decompose the residual $E$ into \textit{duration} and \textit{other} components. The proportion of variance of $E_{dur}$ is .31, and that of $E_{other}$ is .25, indicating that a relatively large amount of information in the residual still concerns duration. The $R^2$ of duration regression using $E_{dur}$ is .28, and that for $E_{other}$ is .10. These numbers are consistent with prosodic factors remaining important in the spectrogram features.


\section{Frobenius norm decomposition under LDL}
\label{appendix-Frobenius}
We use the Frobenius norm to evaluate alignment in Linear Discriminative Learning (LDL), and interpret it as \textit{variance explained}, in a way comparable to $R^2$. This interpretation requires clarification, since \textit{variance} is not formally defined for matrices and because the corresponding decomposition into explained and residual components is not explicit in LDL. This section provides the justification for using the Frobenius norm under this interpretation.

In what follows, we first show how the Frobenius norm is related to the matrix trace and why it is equivalent to the sum of squared matrix elements. Secondly, we show that under LDL, the Frobenius norms of the prediction and the residual can be linearly decomposed.

To begin, we write $C$ as vertically stacked row vectors, and following the convention in LDL, we denote each row vector by \(\bm{c}_i\): 

\[
C =
\begin{pmatrix}
\rule[0.5ex]{0.8cm}{0.4pt}\,\bm{c}_1\,\rule[0.5ex]{0.8cm}{0.4pt}\\
\rule[0.5ex]{0.8cm}{0.4pt}\,\bm{c}_2\,\rule[0.5ex]{0.8cm}{0.4pt}\\
\vdots\\
\rule[0.5ex]{0.8cm}{0.4pt}\,\bm{c}_n\,\rule[0.5ex]{0.8cm}{0.4pt}
\end{pmatrix}.
\]

\noindent Then

\[
CC^\top
=
\begin{pmatrix}
\bm{c}_1 \bm{c}_1^\top & \bm{c}_1 \bm{c}_2^\top & \cdots & \bm{c}_1 \bm{c}_n^\top \\
\bm{c}_2 \bm{c}_1^\top & \bm{c}_2 \bm{c}_2^\top & \cdots & \bm{c}_2 \bm{c}_n^\top \\
\vdots & \vdots & \ddots & \vdots \\
\bm{c}_n \bm{c}_1^\top & \bm{c}_n \bm{c}_2^\top & \cdots & \bm{c}_n \bm{c}_n^\top
\end{pmatrix}.
\]

\noindent The trace picks out the diagonal entries:

\[
\mathrm{tr}(CC^\top)
=
\bm{c}_1 \bm{c}_1^\top
+
\bm{c}_2 \bm{c}_2^\top
+
\cdots
+
\bm{c}_n \bm{c}_n^\top.
\]

\noindent Therefore,

\[
\lVert C \rVert_F^2
=
\sum_{i=1}^n \bm{c}_i \bm{c}_i^\top
=
\mathrm{tr}(CC^\top)
=
\mathrm{tr}(C^\top C).
\]

\noindent The last equality follows from the cyclic property of trace.

In the LDL production route, and similarly in the comprehension route, we
estimate the $G$ matrix by least squares, where speech vectors are predicted
from their corresponding semantic embeddings. That is,

\begin{align*}
C &= SG + (C-SG) \\
  &= \hat{C} + E.
\end{align*}

\noindent The Frobenius norm of $C$ becomes,

\begin{align*}
\lVert C \rVert_F^2
&=
\lVert \hat{C} + E \rVert_F^2 \\
&=
\lVert \hat{C} \rVert_F^2
+
\lVert E \rVert_F^2
+
2\,\mathrm{tr}(\hat{C}^\top E).
\end{align*}

There is a cross-product term in the equation, which equals zero because the least-squares solution decomposes $C$ into two orthogonal components. To see this, we can write $\hat{C}$ in terms of $C$ and a projection matrix $P_S$:

\begin{align*}
\hat{C} & = S (S^\top S)^{-1} S^\top C \\
        & = P_S C.
\end{align*}

The derivation becomes clearer if we write \(S\) using the singular value decomposition as \(U_S \Sigma_S V_S^\top\), and the predicted \(\hat{C}\) can be written as

\begin{align*}
\hat{C} &= P_S C \\
        &= U_S \Sigma_S V_S^\top
           (V_S \Sigma_S^\top \Sigma_S V_S^\top)^{-1}
           V_S \Sigma_S^\top U_S^\top C \\
        &= U_S U_S^\top C.
\end{align*}

Here, since $U_S$ has orthonormal columns, $U_S^\top U_S = I$, which allows us to derive the orthogonality between $\hat{C}$ and $E=C-\hat{C}$:

\begin{align*}
\hat{C}^\top E
&= \hat{C}^\top (C-\hat{C})\\
&= (U_S U_S^\top C)^\top (C-U_S U_S^\top C) \\
&= C^\top U_S U_S^\top (I-U_S U_S^\top) C \\
&= C^\top (U_S U_S^\top - U_S U_S^\top U_S U_S^\top) C \\
&= C^\top (U_S U_S^\top - U_S U_S^\top) C \\
&= 0.
\end{align*}

Since $\hat{C}$ and $E$ are orthogonal, $\mathrm{tr}(\hat{C}^\top E)$ equals zero. The Frobenius norm decomposition of $C$ can therefore be written as

$$
\lVert C \rVert_F^2
=
\lVert \hat{C} \rVert_F^2
+
\lVert E \rVert_F^2,
$$

\noindent which means that the Frobenius norm of $C$ can be expressed as the sum of the Frobenius norms of the prediction $\hat{C}$ and the residual $E$.


\newpage
\bibliography{reference}

\end{document}